\documentclass{article}

\usepackage[main, final]{neurips_2026}

\usepackage[utf8]{inputenc} 
\usepackage[T1]{fontenc}    
\usepackage{hyperref}       
\usepackage{url}            
\usepackage{booktabs}       
\usepackage{amsfonts}       
\usepackage{nicefrac}       
\usepackage{microtype}      
\usepackage{xcolor}         
\usepackage{natbib}

\usepackage{amsmath}
\usepackage{multirow, makecell}
\usepackage{multirow}
\usepackage[dvipsnames]{xcolor}
\usepackage{graphicx}
\usepackage{caption}
\usepackage{makecell}

\usepackage{amssymb}
\usepackage{mathtools}
\usepackage{amsthm}

\theoremstyle{plain}
\newtheorem{theorem}{Theorem}[section]
\newtheorem{proposition}[theorem]{Proposition}
\newtheorem{lemma}[theorem]{Lemma}

\theoremstyle{definition}
\newtheorem{definition}[theorem]{Definition}

\theoremstyle{remark}

\title{Unrolled gradients in disguise: bridging interpolation-based and Jacobian regularization for stable neural dynamics }

\author{%
  Maya Janvier \\
  LOCEAN\\
  Sorbonne University\\
  Paris, France \\
  \texttt{maya.janvier@locean.ipsl.fr} \\
  \And
  Etienne Meunier \\
  Inria \\
  Paris, France \\
  \texttt{etienne.meunier@inria.fr} \\
}

\begin{document}

\maketitle

\begin{abstract}
    Hybrid models and neural differential equations (NDEs) are becoming increasingly important for modeling physical systems; however, they often encounter stability and accuracy issues during long-term integration. Training on unrolled trajectories is known to limit these divergences but quickly becomes too computationally expensive due to the need to backpropagate through the solver, making this method impractical for large-scale physical systems (e.g. computational fluid mechanics, weather and climate forecasting). Two popular paradigms for improving the stability of learned dynamics are to train on interpolated or noisy samples or to regularize the Jacobian of the learned representation. In this work, we demonstrate that these paradigms optimize equivalent objectives as they both correspond to an approximation of the gradients obtained in an unrolled learning strategy. Building on this understanding, we derive a principled framework to regularize training and demonstrate its impact on the stability of learned dynamical systems in both supervised and unsupervised settings.
\end{abstract}

\section{Introduction}
\label{sec:intro}

Autoregressive models are ubiquitous in the deep learning literature and are central tools in many applications, from forecasting \cite{lippe2023pderefinerachievingaccuratelong, lam2023graphcastlearningskillfulmediumrange, yin2023continuouspdedynamicsforecasting, pedersen2025thermalizerstableautoregressiveneural, takamoto2024pdebenchextensivebenchmarkscientific} to video analysis \cite{zhang2025videoincontextlearningautoregressive, li2025learningnextframepredictionautoregressive}. Autoregressive models typically perform two distinct tasks at once: learning an underlying dynamic and learning a solver to advance this ODE in time. These two tasks are intertwined in a way that we cannot decipher the contribution of one or the other. Neural ODE models \cite{chen2019neuralordinarydifferentialequations}, on the other hand, leave the solving part to numerical integration schemes, while a neural network (NN) is used as the dynamical function or right-hand side of the ODE. This separation has many benefits as it provides a more interpretable and flexible representation while encompassing the literature of autoregressive models (recovering ResNets when using Euler forward). It allows for example to use different timesteps at inference, while most autoregressive models can only predict a fixed number of timesteps ahead. Neural ODEs gained popularity across a wide range of deep learning tasks, from the modeling of physical systems \cite{white2024stabilizedneuraldifferentialequations, Yin_2021, Linot_2023, pal2025semiexplicitneuraldaeslearning}, to classification tasks \cite{yan2022robustnessneuralordinarydifferential, kang2021stableneuralodelyapunovstable} and generative tasks \cite{finlay2020trainneuralodeworld, song2021scorebasedgenerativemodelingstochastic}. There is a particular and renewed interest in them for climate and weather forecasting as they provide a natural framework for hybrid modeling \cite{bolibar:hal-05506928, meunier2025fullydifferentiableneuralocean, Kochkov_2024}. 

There are two settings for the training of Neural ODEs: the supervised case, where we have access to the true dynamic, and the unsupervised case where this dynamic is unknown and we only have access to trajectories. The former is interesting for building fast emulators of models (to accelerate General Circulation Models predictions where it takes months to perform a hundred years of simulations for example) or in hybrid models where the learned dynamic is tasked only to correct the biases of the existing one (parameterizing high resolution phenomena in low resolution models \cite{Thiry_2022}). The unsupervised case on the other hand corresponds to more realistic scenarios (observed trajectories). 

In the supervised case, the Neural ODE can be directly trained to match the right-hand side dynamic in what is called offline or a-priori training \cite{frezat2021posteriorilearningquasigeostrophicturbulence}. However, a core challenge that arises is that such models quickly become unstable at inference when integrated over long-rollouts \cite{frezat2021posteriorilearningquasigeostrophicturbulence, takamoto2024pdebenchextensivebenchmarkscientific}. Training on unrolled sequences through the solver (online or a-posteriori training) is known to mitigate this problem and no longer requires the knowledge of $F$ (unsupervised case); however, this method quickly becomes too computationally expensive due to the need to compute gradients over an iterative process while also suffering from an explosion of the gradients during unrolling \cite{list2024differentiabilityunrolledtrainingneural}. Although methods have been explored to mitigate the latter \cite{hess2023, mikhaeil2022difficultylearningchaoticdynamics}, the issue of the cost remains prominent. On real-world, large scale applications that involve timescales of months to years (weather forecasting), the cost of long rollouts becomes entirely prohibitive while offline training fails in producing stable models for long-horizon predictions \cite{bolton2019, zanna2020, ross2023, pedersen2023reliablecoarsegrainedturbulentsimulations}. This limits Neural ODEs usage to simple dynamical systems in practice, such as robotic simulations \cite{robotics2024, white2024stabilizedneuraldifferentialequations}. 

Among the various methods developed to improve stability (see Section \ref{sec:related}), the approaches that can be applied in an offline setting are thus the preferred solutions to this challenge. In particular, methods that either regularize the Jacobian of the system \cite{nie2026jawsenhancinglongtermrollout, xie2024smoothsparselatentdynamics} or train on interpolated or noisy samples \cite{brandstetter2023messagepassingneuralpde, hou2025cfolearningcontinuoustimepde} give promising results on complex physical systems, including chaotic ones. In this paper, we show that these paradigms are not inherently different but rather correspond to an approximation of the gradients in the state-of-the-art online training (Section \ref{sec:maths}). We then exploit this common framework to derive principled regularization loss functions (Section \ref{sec:method}) and investigate their efficiency on various physical problems (Section \ref{sec:results}) in both supervised and unsupervised settings. 

\section{Related work}
\label{sec:related}

\subsection{One challenge, various strategies}
We identify four categories of approaches to mitigate the stability issue of Neural ODEs. 

\paragraph{Projection and energy conservation.} Early attempts at stabilizing Neural ODEs focused on energy conservation by introducing dissipation  \cite{Linot_2023}. Projecting onto an underlying manifold (e.g., algebraic constraints) allows stabilizing long-term dynamics learned on short rollouts \cite{white2024stabilizedneuraldifferentialequations, pal2025semiexplicitneuraldaeslearning} and thus alleviating the computational cost of unrolling. However, these methods are inherently limited to conservative systems and projections are prohibitively expansive for large-scale systems. Autoregressive models trained to correct the accumulation of errors in the form of denoising diffusion processes are efficient \cite{lippe2023pderefinerachievingaccuratelong, pedersen2025thermalizerstableautoregressiveneural}, but can be cumbersome to train and rely on a number of denoising steps, which is a sensitive and hard parameter to tune in order not to break the temporal coherence of the predicted trajectories. 

\paragraph{Controlling the Lipschitz constant.} Stability in autoregressive models is closely linked to robustness against adversarial perturbations, that is to say ensuring that similar inputs (for a certain norm) produce similar outputs. The Lipschitz constant encompasses this property (annex \ref{a:def}) and can be leveraged to enforce robustness. Unfortunately, estimating a computationally efficient Lipschitz constant for training is an open research question \cite{raghunathan2018semidefiniterelaxationscertifyingrobustness, 
scaman2019lipschitzregularitydeepneural, jordan2021exactlycomputinglocallipschitz, 
huang2021trainingcertifiablyrobustneural,
Pauli_2022,
fazlyab2023efficientaccurateestimationlipschitz,
pintore2024computablelipschitzboundsdeep}. Hard constraints in the form of Lipschitz layers \cite{meunier2022dynamicalperspectivelipschitzneural, wang2023directparameterizationlipschitzboundeddeep, huang2021trainingcertifiablyrobustneural}, energy conserving architectures \cite{revay2023recurrentequilibriumnetworksflexible} or penalizations to ensure a contracting Lipschitz constant ($L<1$) \cite{yan2022robustnessneuralordinarydifferential} directly enforce robust predictions of steady states and stationarity, which is no longer valid when it comes to model chaotic dynamics that are constantly evolving and never reach steady states. Furthermore, hard constraints limit the expressiveness of the model which may harm its performances and generalization capacities. 

\paragraph{Jacobian regularization.} Softer constraints focusing on the Jacobian (annex \ref{a:def}) have been deployed, such as spatially-adaptive Jacobian regularization in the real space (JAWS) \cite{nie2026jawsenhancinglongtermrollout} or third-order time derivative penalization obtained by finite differences in the latent space (Jerk) \cite{xie2024smoothsparselatentdynamics}, however these methods still need online training to really improve stability. Spectral constraints acting on the eigenvalues of the Jacobian can ensure stability around equilibrium points \cite{kang2021stableneuralodelyapunovstable} or within integration schemes \cite{westny2024stabilityinformedinitializationneuralordinary} (A-stability) and are known to stabilize training \cite{tomihari2025recurrentselfattentiondynamicsenergyagnostic}, accelerate convergence \cite{novak2018sensitivitygeneralizationneuralnetworks, finlay2020trainneuralodeworld, kelly2020learningdifferentialequationseasy} and promote continual training of models by maintaining gradient diversity throughout training \cite{lewandowski2024learningcontinuallyspectralregularization}.These observations are qualitative and made on a limited set of physical systems, the lack of theoretical understanding limits their generalizability. 

\paragraph{Data augmentation and interpolation.} Robustness can also be achieved through adversarial training on noisy samples \cite{sanchezgonzalez2020learningsimulatecomplexphysics, mayr2023boundarygraphneuralnetworks}. Such methods would also result in a contracting representation (proof in annex \ref{a:noisy}), limiting their efficiency in modeling chaotic physical systems. The pushforward trick \cite{brandstetter2023messagepassingneuralpde} achieves a form of noisy training by unrolling for two steps and backpropagating gradients only through the last step, but its efficiency still lacks theoretical understanding. Continuous Flow Operators \cite{hou2025cfolearningcontinuoustimepde} (CFO)
fit temporal splines using finite-difference estimates of time derivatives at true trajectory points to closely approximate the true PDE dynamics and train in an offline fashion on new data points along the spline. 

\subsection{Our work: bridging interpolation-based and Jacobian regularization.}
Overall, the energy and Lipschitz approaches often are system-dependent and lack flexibility. Jacobian regularization and interpolation approaches on the other hand are more systemic/general methods in tackling the issue of stability in Neural ODEs, but still lack theoretical understanding. In this paper, we bridge this understanding gap, focusing on Jacobian regularization (JAWS) and interpolation (CFO) methods as they have a particular advantage: both can be performed not only in an offline fashion (that is, the training is performed outside the solver) but also without having access to the true dynamic function $F$: the former leverages the learned Jacobian $J_{F_\theta}$; the latter constructs probability paths whose velocities closely approximate the true $F$, also relying on finite differences estimation. We will: (i) prove that these approaches can be understood as regularization of gradients obtained in an unrolled learning strategy (involving the Jacobian in particular), (ii) derive a principled framework to investigate all of the terms that appear from unrolling, and (iii) analyze its efficiency on different physical systems in both supervised and unsupervised settings.

\section{Unrolled gradients in disguise: on the importance of the Jacobian}
\label{sec:maths}

Let us consider a general Partial Differential Equation (or PDE) in the form of 
\begin{equation}
    \frac{\partial \textbf{x}}{\partial t}=F(t, \textbf{x}, \nabla \textbf{x}, \nabla ^2 \textbf{x}, ...)
\end{equation}
with \textbf{x} the prognostic variables evolving temporally through the dynamic function $F$. To simulate trajectories of \textbf{x}, we numerically integrate the PDE forward in time using a temporal integration scheme or solver. Training a neural network $F_\theta$ to approximate $F$ from observed trajectories is a central task in scientific ML. The length of the trajectories we feed to the neural network, measured in number of integration steps performed through the solver, is called the training rollout $N$. Since non-autonomous systems described by $\frac{dx}{dt}=F(x(t), t)$ can always be rewritten as autonomous ones by using the augmented state $X= (x(t), t)$ and $\frac{dX}{dt}= (F(X), 1)$, we will use the autonomous formulation in the paper without losing generality. 

\subsection{General influence of the Lipschitz constant}
Neural ODEs \cite{chen2019neuralordinarydifferentialequations} combine data-driven estimators and numerical solvers. Consider a neural network $F_\theta$ representing the right-hand side or dynamics of an autonomous ODE $\frac{dx_\theta(t)}{dt}=F_\theta(x_\theta(t))$, associated with an integration scheme (or solver) and a step-size $\Delta t$, that aims to produce the same trajectories as the true ODE $\frac{dx(t)}{dt}=F(x(t))$, namely $X \in \mathcal{T}_F=\{x_i=x_0+\int_{t_0}^{t_i}F(x(s))ds, 0 \leq i \leq N\}$. We consider both $F$ and $F_\theta$ to be differentiable and Lipschitz continuous. 

In offline training, the model is tasked to minimize $\Vert F(x)-F_\theta(x)\Vert_2^2$, so that learning $F$ is consequently achieved in a direct way, without using the solver.  However, this method needs access to the true $F$ which is unknown in most scenarios. Furthermore, when applied to real world problems where the true dynamic is approximated by a high resolution simulation, this method leads to diverging trajectories for long-term integration at inference \cite{frezat2021posteriorilearningquasigeostrophicturbulence, list2024differentiabilityunrolledtrainingneural}. Looking at bounds on the trajectory error \cite{Yin_2021,yan2022robustnessneuralordinarydifferential}, it is easy to explain the failure of offline learning. Using Grönwall's lemma \ref{l:gronwall}, we get for the trajectory error starting from the same initial conditions (proof in appendix \ref{a:lipschitz}):  
\begin{equation}
\label{eq:gronwall}
    \forall t \geq0, \,\Vert x(t)-x_\theta(t)\Vert_2 \leq \frac{\Vert\epsilon_\theta\Vert_\infty}{L_{F_\theta}} (\exp (L_{F_\theta}t)-1)
\end{equation}
with $\epsilon_\theta=F-F_\theta$ the offline error and $L_{F_\theta}$ the Lipschitz constant of $F_\theta$. Thus, a model with a low $\Vert\epsilon_\theta\Vert_\infty$ (minimized in offline training) but a large $L_{F_\theta}$ can still lead to diverging trajectories as $t$ grows, driven by the exponential term in the upper bound. This bound shows that there is a trade-off between the smoothness of a neural network and its expressiveness, i.e., between $\epsilon_\theta$ and $L_{F_\theta}$, that is not trivial to achieve in practice \cite{huang2021trainingcertifiablyrobustneural, wang2022quantitativegeometricapproachneuralnetwork, fazlyab2023efficientaccurateestimationlipschitz}. 

On the other hand, performing the training via unrolled trajectories (or online learning) only requires access to trajectory points, and no longer to $F$ (unsupervised case). It is achieved by minimizing the trajectory loss \cite{chen2019neuralordinarydifferentialequations, Yin_2021, Linot_2023}:
\begin{equation}
\label{eq:traj_loss}
    \mathcal{L}^N_{\text{traj}}(\theta)=\sum_{X\in \mathcal{T}_{F}}\sum_{i=0}^N\Vert x(t_i)-\tilde{x}_\theta(t_i)\Vert_2^2  \text{,} \quad t_i=i\Delta t
\end{equation}
with $\tilde{x}_\theta(t_i)$ obtained by numerically integrating the ODE forward in time for $0 \leq i \leq N$ steps. As seen in the introduction, unrolling for long enough sequences results in much more stable rollouts \cite{ frezat2021posteriorilearningquasigeostrophicturbulence, list2024differentiabilityunrolledtrainingneural}. However, there is still little theoretical understanding of the reasons behind this property. This is what we investigate in the next section. 

\subsection{What is actually hidden in unrolled trajectories ?}

Examining the terms that arise during unrolling, we better understand where the stability of online training stems from. We use the Taylor-Young formula \ref{t:taylor} to linearize around the data points and see the different terms at play. Starting from some $x(t)=x_t$ (and similarly for $x_\theta(t)=x_t^\theta$), we derive the trajectory error for $N=2$ steps under Euler forward (full proof in annex \ref{a:unrolling}):
\begin{align}
\label{eq:unrolling}
x_{t+2\Delta t} - x^\theta_{t+2\Delta t}
&= \underbrace{(x_t - x^\theta_t)}_{\text{initial error}} \\
&\quad + 2\,\Delta t\, [\underbrace{\textcolor{RoyalBlue}{\epsilon_\theta(x_t)}}_{\text{offline error}} + \underbrace{\textcolor{ForestGreen}{J_{F_\theta}(x_t)}\cdot(x_t - x^\theta_t)}_{\text{Grönwall drift}}\, + \quad o(\Vert x_t-x_t^\theta\Vert_2)] \notag\\
&\quad + \Delta t^2\, [\underbrace{\textcolor{Red}{J_{\epsilon_\theta}(x_t)\cdot F(x_t)}}_{\text{Jacobian match in true direction}} \quad 
+  \underbrace{\textcolor{ForestGreen}{J_{F_\theta}(x_t)}\cdot \textcolor{RoyalBlue}{\epsilon_\theta(x_t)}}_{\text{Lipschitz stretch in direction of error}} \notag
\quad \notag \\
&\quad + \underbrace{\textcolor{ForestGreen}{J_{F_\theta}(x_t)}\cdot (F_\theta(x_t) - F_\theta(x^\theta_t))}_{\text{Grönwall drift on derivatives}} + \quad o(\Vert x_t-x_t^\theta\Vert_2)]\quad \notag \\
&\quad + \, o(\Delta t^2)
\end{align}
In this formula, we first recover the \textcolor{RoyalBlue}{offline error} and the Lipschitz contribution from the derivation of the Grönwall bounds (annex \ref{a:lipschitz}), terms in $O(\Delta t)$ that appear naturally with the first unrolling step ($t$ to $t+\Delta t$). Unrolling for a second step gives terms in $o(\Delta t)$ that involve directional derivatives of the true and the learned Jacobians in specific directions. Equation \ref{eq:unrolling} shows that even if we train the model with the trajectory loss by unrolling until $t$, the error for a longer inference (up to $t+2\Delta t$ in eq. \ref{eq:unrolling}) will involve those terms that need to be appropriately learned, which is a new insight on the trajectory loss. 

In particular, we now understand why regularizing the norm of the \textcolor{ForestGreen}{Jacobian of the learned dynamic $J_{F_\theta}$} as done in JAWS (annex \ref{sec:jaws}) is efficient in stabilizing the long-term dynamics: controlling it limits the scale of the propagation of initial errors. However, regularizing the learned Jacobian does not capture the \textcolor{Red}{Jacobian matching term $J_{\epsilon_\theta}(x_t) \cdot F(x_t)$}$=(J_F(x_t)-J_{F_\theta}(x_t))\cdot F(x_t)$ ensuring that $J_F$ and $J_{F_\theta}$ resemble each other in realistic directions of the trajectories: it can explain why in practice JAWS still need short-online training ($N=5$) to successfully stabilize long-term dynamics.

We now analyze equation \ref{eq:unrolling} in the training regime, i.e., when the trajectories have the same initial conditions $x_t=x^\theta_t$:
\begin{align}
\label{eq:n2-1}
x_{t+2\Delta t} - x^\theta_{t+2\Delta t}
&= 2\,\Delta t\ \textcolor{RoyalBlue}{\epsilon_\theta(x_t)} + \Delta t^2\, (\textcolor{Red}{J_{\epsilon_\theta}(x_t)\cdot F(x_t)}
+  J_{F_\theta}(x_t) \cdot\epsilon_\theta(x_t)) + o(\Delta t^2)\\
\label{eq:n2-2}
&= 2\,\Delta t\ \textcolor{RoyalBlue}{\epsilon_\theta(x_t)} + \Delta t^2 (\underbrace{\textcolor{Brown}{J_{F}(x_t)\cdot F(x_t)}}_{d^2x/dt^2} \textcolor{Brown}{- J_{F_\theta}(x_t)\cdot F_\theta(x_t)})  + o(\Delta t^2)
\end{align}
This equation shows that the unrolled loss, i.e., the trajectory loss $N=2$ can be approximated, accurate to $o(\Delta t^2$), by an expression involving not only the \textcolor{RoyalBlue}{offline error $\epsilon_\theta$} but also another offline term, the \textcolor{Brown}{acceleration error $\eta_\theta(x_t)$}$ := J_F(x_t) \cdot F(x_t) -J_{F_\theta}(x_t) \cdot F_{\theta}(x_t)$ involving the Jacobians of the systems. This formula can be generalized to $N\geq 2$ steps and to other solvers such as RK2 (further detailed in annex \ref{a:unrolling}). Under sufficient regularity assumptions on $F$, $F_\theta$ and their derivatives, the gradients of the approximation terms in eq. \ref{eq:n2-2} (right side) with respect to the parameter $\theta$ remain close to the online gradients (left side), with a relative error of $O(\Delta t^2)$ (proof in \ref{a:unroll_grad}). The benefit is that this approximation comes at much lower cost than doing direct online training since it does not backpropagate through the auto-regressive steps and doesn’t require a differentiable solver.

In addition, we found that CFO is actually acting on the offline error and Jacobian matching term during training by linearizing its loss: 
\begin{equation}
    \mathcal{L}_{\text{CFO}}(\theta)=\sum_{X\in \mathcal{T}_{\text{I}}} \Vert F_\theta(x(t_i))-\partial_tI(t_i,x(t_i))\Vert^2_2
\end{equation}
where $I$ is the fitted spline and $\mathcal{T}_{\text{I}}$ the ensemble of data points sampled along $I$. In the linear spline fitting, CFO consists in sampling new points on the line between $x_t$ and $x_{t+\Delta t}$: $\text{for } \gamma \in [0, \Delta t], x_t^\gamma= x_t + \gamma F(x_t)$. If the fit was perfect or if we had access to the true $F$, the CFO loss on linearly interpolated points would yield with a first order Taylor-Young approximation:
\begin{align}
    F(x_t^\gamma)-F_\theta(x_t^\gamma) &=F(x_t+\gamma F(x_t))- F_\theta(x_t + \gamma F(x_t)) \\
    &= F(x_t) + \gamma J_F(x_t)\cdot F(x_t) - F_\theta(x_t) - \gamma J_{F_\theta}(x_t)\cdot F(x_t) + o(\Delta t)\\
    \label{eq:cfo-lin}
    &= \underbrace{\textcolor{RoyalBlue}{\epsilon_\theta(x_t)}}_{\text{offline error}} + \gamma \underbrace{\textcolor{Red}{J_{\epsilon_\theta}(x_t) \cdot F(x_t)}}_{\text{Jacobian match in true directions}} + o(\Delta t)
\end{align}

As in equation \ref{eq:n2-1}, the CFO loss also constrains the offline error and the \textcolor{Red}{Jacobian matching term} (eq. \ref{eq:cfo-lin}). CFO sampling acts partly as unrolling by characterizing the Jacobians: it explains its success in long-horizon stability. It also explains the stabilization and acceleration of the training process observed in \cite{hou2025cfolearningcontinuoustimepde} as it is a known consequence of a better characterization of the Jacobian documented in
\cite{novak2018sensitivitygeneralizationneuralnetworks, finlay2020trainneuralodeworld, kelly2020learningdifferentialequationseasy,
lewandowski2024learningcontinuallyspectralregularization,
tomihari2025recurrentselfattentiondynamicsenergyagnostic}. Contrary to JAWS, the norm of \textcolor{ForestGreen}{$J_{F_\theta}$} is not constrained at all in the CFO formulation, and thus not controlling the complete term involved when unrolling: the acceleration error $J_{F}(x_t) \cdot F(x_t) - J_{F_\theta}(x_t) \cdot F_\theta(x_t)$ in equation \ref{eq:n2-2}.

In the end, we see that the efficiency of both Jacobian regularization such as JAWS and interpolation-based strategies such as CFO is rooted in the different Jacobian characterizations that appear when linearizing the unrolled trajectory loss and that are simply absent in offline training. This derivation also exhibits offline approximations of the online trajectory loss, in particular the acceleration matching formulation in eq. \ref{eq:n2-2}, which we leverage in the next section either as a training objective in an offline setting or regularization term in an online one. 

\section{Methods}
\label{sec:method}

It comes naturally from the previous section that minimizing the terms at play in equation \ref{eq:unrolling} should improve long-term stability. We focus on two challenging cases: \textbf{(1) the offline setting}, where we directly use the offline formulations of the losses as described in equations \ref{eq:n2-2} and \ref{eq:cfo-lin}, and \textbf{(2) the short-rollout training} where the model is trained online on short rollouts ($N \leq 5$) with regularization terms derived from equation \ref{eq:unrolling}.

\subsection{Offline learning via spline fitting}
\label{sec:interp}

From equations \ref{eq:n2-2} and \ref{eq:cfo-lin}, we can derive offline versions of AccMatch and JacMatch to compare to plain offline training, with $\gamma$ a hyperparameter that we tune on a held out validation set. We can evaluate them in both supervised and unsupervised settings, by using either the true $F$ or the spline fitting technique of CFO. Compared to using finite differences approximations, this approach avoids the noise amplification issue (when $\Delta t \rightarrow 0$, more in next section) by directly fitting a cubic spline to $F$ from trajectory points, inducing a smooth representation at the cost of a less accurate one. 

\begin{figure*}[t]  
  \centering
  \includegraphics[width=0.51\textwidth]{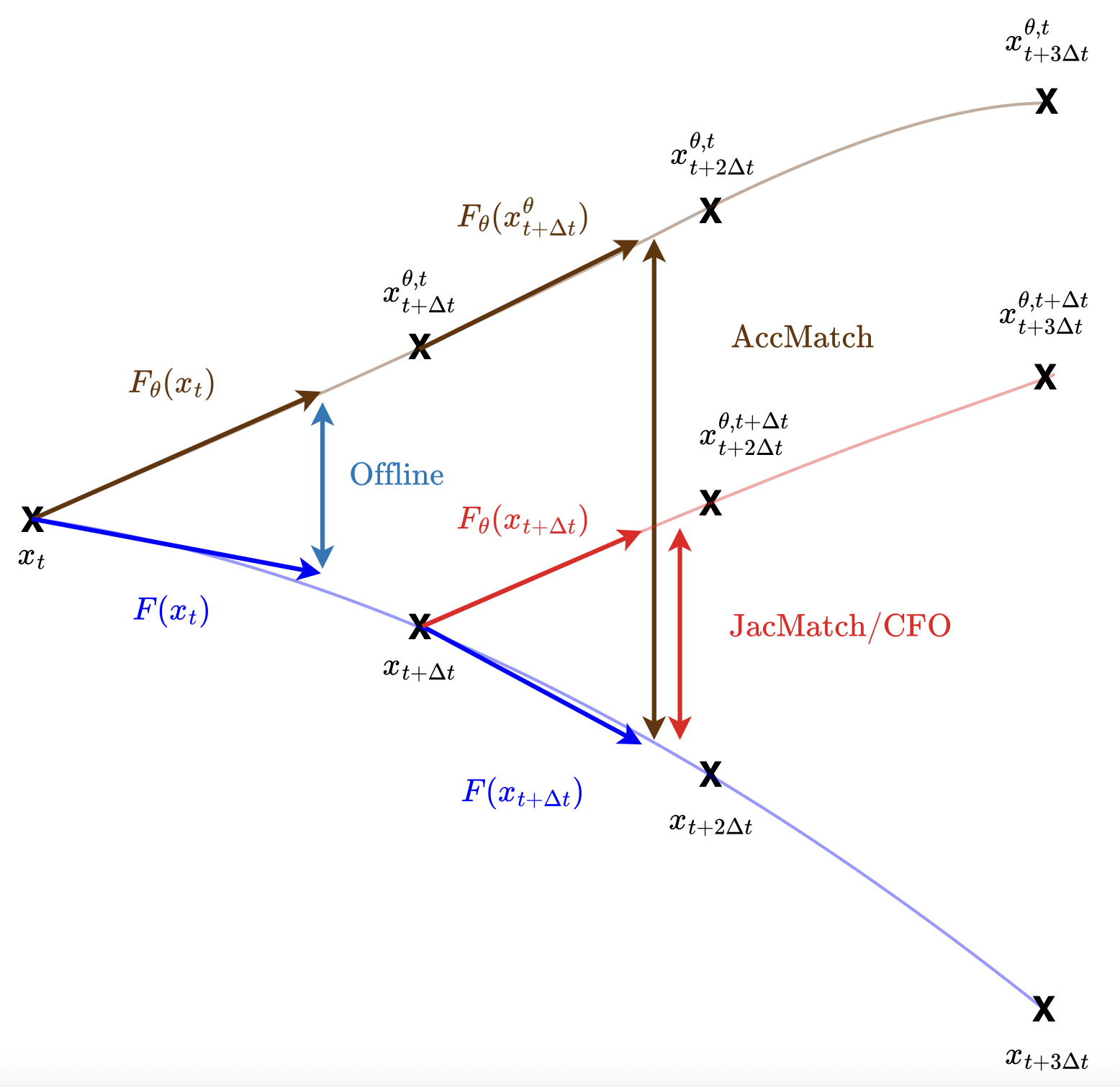}
  \caption{\textbf{Losses interpretation.} When $\gamma=1$, $x_t^\gamma=x_{t+\Delta t}$ and we get that JacMatch is equivalent to the error of a delayed $F_\theta$ trajectory starting from $x_{t+\Delta t}$, i.e., only accounting for the error between the Jacobians while AccMatch also accounts for the error after the first timestep.}
  \label{fig:losses_scheme}
\end{figure*}

We summarize the offline methods we will test in Table \ref{tab:methods}. To give a better intuition of what the losses represent, Figure \ref{fig:losses_scheme} illustrates the discrepancies they target in an unrolled trajectory: offline acts on the direction divergence the trajectories $x$ and $x^\theta$ take in the first step from the same initial conditions, while AccMatch looks at reducing this direction gap after a second step. JacMatch and CFO, considering $\gamma=1$, also target a single step divergence, but starting from the later $x_{t+\Delta t}$ state.

\begin{table}[t]
\caption{\textbf{Summary of the methods}. \textbf{(i) Offline :} The original offline method only uses the exact $F$ and CFO the fitted spline version of it. CFO differs from JacMatch by the upsampling (10 more $x_t^\gamma$ data points), although they come from the same theoretical formula. \textbf{(ii) Short-online:} the regularizations are tested within the training loss: $\mathcal{L}(\theta)= \mathcal{L}_{\text{traj}}^N(\theta) + \lambda \mathcal{L}_{\text{reg}}(\theta)$. The Lipschitz method refers to the Jacobian regularization approach, in the spirit of JAWS. To build an unsupervised version of JacMatch, we use a first-order finite differences (FD) approximation (eq. \ref{eq:jac_unsup}). A FD version of AccMatch is possible but not tested here. \textbf{Both:} the $\gamma$ and $\lambda$ hyperparameters are tuned on a held-out validation set. }
\label{tab:methods}
  \begin{center}
    \begin{small}
      \begin{sc}
    \begin{tabular}{|l|c|c|c|}
    \hline
    \textbf{Method} & \textbf{Exact form} & \textbf{\makecell{Linearized form \\ (offline)}} & \textbf{\makecell{Regularization form \\ (short-online)}} \\
    \hline
    
    \textcolor{RoyalBlue}{Offline} & $ \Vert F(x_t)-F_\theta(x_t) \Vert_2^2$ & \makecell{$\Vert \epsilon_\theta(x_t) \Vert_2^2$ \\ exact $F$} & X \\

     & & & \\

    \makecell{\textcolor{Green}{Lipschitz}}  & $\Vert J_{F_\theta}(x_t)\Vert_F^2$ & X & $\frac{1}{V} \sum_{j=1}^V \Vert J_{F_\theta}(x_{t}) \cdot v_j \Vert_2^2$ \\ 

    & & & \\

    CFO & $\Vert F(x_{t}^\gamma)-F_\theta(x_{t}^\gamma) \Vert_2^2$ & \makecell{$\Vert F(x_{t}^\gamma)-F_\theta(x_{t}^\gamma) \Vert_2^2$ \\
    spline fit $F$ \\ various $\gamma$ (upsampling x10)} & X \\

    & & & \\

    \textcolor{Red}{JacMatch} &$\Vert F(x_{t}^\gamma)-F_\theta(x_{t}^\gamma)\Vert_2^2$ & \makecell{$\Vert \epsilon_\theta(x_t) + \gamma J_{\epsilon_\theta}(x_t) \cdot F(x_t)  \Vert_2^2$ \\
    exact or spline fit $F$, $J_F$\\ single $\gamma$} & \makecell{$\Vert  J_{\epsilon_\theta}(x_t) \cdot F(x_t) \Vert_2^2$\\ exact or FD $F$, $J_F$ [\ref{eq:jac_unsup}] \\ single $\lambda$}  \\

    & & & \\
    
    \textcolor{Brown}{AccMatch} &$\Vert F(x_{t}^\gamma)-F_\theta(x_{t}^{\theta,\gamma}) \Vert_2^2$ & \makecell{$\Vert \epsilon_\theta(x_t) + \gamma \eta_\theta(x_t) \Vert_2^2$ \\
    exact or spline fit $F$, $J_F$  \\ single $\gamma$} & \makecell{$\Vert \eta_\theta(x_t) \Vert_2^2$ \\ exact (or FD) $F$, $J_F$  \\ single $\lambda$}  \\

    \hline
    \end{tabular}
\end{sc}
\end{small}
\end{center}
\end{table}

\subsection{Regularizing short-online training}
\label{sec:online}

The goal of this experiment is to study, understand and compare the limitations and efficiencies of this family of losses when they are used as regularizations in the following training loss:
\begin{equation}
    \mathcal{L}(\theta)= \mathcal{L}_{\text{traj}}^N(\theta) + \lambda \mathcal{L}_{\text{reg}}(\theta)
\end{equation}
where $N$ is short-term to avoid prohibitive computational costs, and $\lambda$ a hyperparameter that we tune on a held out validation set. This extends the offline methods to situations where short-online training is possible, under both supervised and unsupervised settings. The regularizations are summarized in Table \ref{tab:methods}. 

\subsubsection{Supervised approaches}

From the \textcolor{Red}{Jacobian matching term} in equation \ref{eq:unrolling}, we derive: 
    \begin{equation}
        \mathcal{L}_{\text{Jacmatch}}^{sup}(\theta)=\sum_{X\in \mathcal{T}_F} \sum_{i=0}^N \Vert (J_{F}(x_{t_i})- J_{F_\theta}(x_{t_i})) \cdot F(x_{t_i}) \Vert_2^2
    \end{equation} 
From equation \ref{eq:n2-2}, we define the \textcolor{Brown}{acceleration matching term}:
    \begin{equation}
        \mathcal{L}_{\text{Accmatch}}^{sup}(\theta) = \sum_{X\in \mathcal{T}_F} \sum_{i=0}^N \Vert J_F(x_{t_i})\cdot F(x_{t_i}) - J_{F_\theta}(x_{t_i}) \cdot {F_\theta}(x_{t_i}) \Vert_2^2
    \end{equation}

$\mathcal{L}_{\text{Jacmatch}}$ and $\mathcal{L}_{\text{Accmatch}}$ are first evaluated in a supervised setting (denoted with $sup$ exponent) where $F$ is accessible, which corresponds to cases where we want to build a fast emulator of the system. This setting represents some gold standard to assess the potential of the different regularizations. We also believe that they will yield similar results in a short-rollout context. Indeed, the difference between the two is only the $J_{F_\theta}(x_t) \cdot \epsilon_\theta(x_t)$ term in equation \ref{eq:n2-1}, and $\epsilon_\theta(x_t)$ will quickly become small for short rollouts $N$ as $\mathcal{L}_\text{traj}^N$ is optimized during training. 

\subsubsection{Unsupervised approaches} 

The \textcolor{ForestGreen}{Jacobian regularization} approach is natively unsupervised as it only requires $J_{F_\theta}$:
    \begin{align}
        \mathcal{L}_{\text{Lip}}(\theta) =\sum_{X\in \mathcal{T}_F} \sum_{i=0}^N \frac{1}{V} \sum_{j=1}^V \Vert J_{F_\theta}(x_{t_i}) \cdot v_j \Vert_2^2 \approx \sum_{X\in \mathcal{T}_F} \sum_{i=0}^N \Vert J_{F_\theta}(x_{t_i}) \Vert_F^2 
    \end{align}
This loss (similar to JAWS, see \ref{sec:jaws}) penalizes the norm of the Jacobian of the learned dynamic, computed via the Hutchinson trace estimator \cite{Hutchinson01011989}. We use $V=10$ directions drawn from $\mathcal{N}(0,I)$ at each epoch, to resemble noise injection strategies (annex \ref{a:noisy}). This Monte-Carlo estimation is unbiased (eq. \ref{eq:hutch}), with its variance decreasing in $\frac{1}{V}$ \cite{Hu_2024}. We found that $V=10$ was a good compromise between computational cost and accuracy in practice.
    
We also derive \textcolor{Purple}{an unsupervised version of JacMatch} using finite differences (FD): 
\begin{equation}
\label{eq:jac_unsup}
    \mathcal{L}_{\text{Jacmatch}}^{unsup}(\theta) =\sum_{X\in \mathcal{T}_F} \sum_{i=0}^{N-2}  \Vert \frac{F_\theta(x_{t_i+\Delta t})-F_\theta(x_{t_i})}{\Delta t} - \frac{x_{t_i + 2\Delta t}-2x_{t_i+ \Delta t}+ x_{t_i}}{\Delta t^2}\Vert_2^2
\end{equation}
The fraction on the right is simply a first order FD approximation of $\frac{d^2x}{dt^2}=J_F(x_t)\cdot F(x_t)$, the acceleration of the true trajectory in $x_t$.
In addition, using the Taylor-Young expansion (\ref{t:taylor}) we have for the left term that: 
\begin{align}
    \frac{F_\theta(x_{t_i+\Delta t}) -F_\theta(x_{t_i})}{\Delta t} &= J_{F_\theta}(x_{t_i}) \cdot\frac{x_{t_i+\Delta t}-x_{t_i} }{\Delta t} + \frac{1}{\Delta t}O((x_{t_i+\Delta t}-x_{t_i})^2) \\
    &= J_{F_\theta}(x_{t_i}) \cdot F(x_{t_i}) + O(\Delta t)
\end{align}
as $x_{t_i+\Delta t}-x_{t_i}=\Delta tF(x_{t_i})+O(\Delta t^2)$. We effectively compute a first order approximation of $J_{F_\theta}(x_t)\cdot F(x_t)$ and we recover an unsupervised version of $\mathcal{L}_{\text{Jacmatch}}$. The bias of the FD approximation could be reduced by using centered stencils in order to get a $O(\Delta t^2)$, while the choice of the FD order is to be balanced with the noise amplification that occurs when $\Delta t \rightarrow 0 $ (caused by $\Delta t$ terms in the denominator).

\section{Results}
\label{sec:results}
Both experimental settings feature training on short rollout/short-horizon trajectories and evaluating on longer ones (annex \ref{a:implementation} for the implementation details). The code is available on GitHub for both the offline and short-online experiments \footnote{\url{https://github.com/mayajanvier/jax-phynity}}. 

\subsection{Offline training}

In this setting, models are trained offline and evaluated on long-term rollouts. Experiments are led on the Van der Pool oscillator: a classical benchmark for Neural ODE approaches presenting a non-linear dynamic with a transient phase before reaching a periodic attractor (annex \ref{a:implementation}). We used 3 different seeds and report the mean long-term MSE accross them here. 

Table \ref{tab:spline} shows that in the plain offline setting, there are no major differences between using the exact $F$ or the fitted spline when the $\Delta t$ of the model is small, however the spline becomes more efficient when it increases. We hypothesise that this potentially smoother version of $F$ may make the training objective easier to optimize and include a native regularization on the smoothness of the learned regularization. This can be linked to previous works \cite{finlay2020trainneuralodeworld, kelly2020learningdifferentialequationseasy} that penalize high-order derivatives to prevent learning overly complex dynamics and encourage smoother paths. 

\begin{table}[h]
    \begin{minipage}[t]{0.35\textwidth}
    \caption{\textbf{Impact of spline fitting}}
    \label{tab:spline}
        \begin{center}
            \begin{small}
                \begin{sc}
                    \begin{tabular}{c|c|c}
            $F$ / $\Delta t$ & 0.01 & 0.1 \\
            \hline
            Exact & 1.354e-4 & 3.556e-4 \\
            Spline & 1.458e-4 & 1.789e-4 \\ 
        \end{tabular}
                \end{sc}
            \end{small}
        \end{center}
        
    \end{minipage}
    \hfill
    \begin{minipage}[t]{0.64\textwidth}
        \caption{\textbf{Long-term MSE (mean accross 3 seeds).} All methods use spline fitting. Best is in bold.}
        \label{tab:offline}
        \begin{center}
            \begin{small}
                \begin{sc}
                    \begin{tabular}{c|c|c|c|c}
            \textbf{Method} & \textcolor{RoyalBlue}{Offline} & \textcolor{Brown}{AccMatch} & \textcolor{Red}{JacMatch} & CFO \\

            \textbf{Score}
            & 1.789e-4
            & \textbf{8.330e-5}
            & 1.180e-4
            & 1.443e-4
            
            \end{tabular}
                \end{sc}
            \end{small}
        \end{center}
        
    \end{minipage}
\end{table}

When using spline fitting, AccMatch performs best, in particular outperforming CFO, while all methods outperform plain offline. CFO and JacMatch perform similarly as expected from eq. \ref{eq:cfo-lin}. We believe that the small difference between the two methods is due to the enhanced identifiability that comes when optimizing on the  explicit Jacobian of the neural network to learn (akin to Hermite interpolation), although the losses’ values will be similar.

\subsection{Short-online training}
We test our regularizations on common one-dimensional ODE and PDE benchmarks treated in other papers that cover a variety of dynamics: a first order non-chaotic autonomous system with Rigid Body \cite{white2024stabilizedneuraldifferentialequations} as well as a chaotic fourth-order partial derivative equation with Kuramoto-Sivashinksy \cite{Linot_2023, lippe2023pderefinerachievingaccuratelong}. Both present either conservation laws or holonomic constraints that should be preserved during integration. As in White \cite{white2024stabilizedneuraldifferentialequations}, we are interested in the relative error between the predicted state $x_\theta(t)$ and the ground truth $x(t)$ over time, averaged over many trajectories with various initial conditions, referred as $R_e(\theta,t)$, as well as temporal evolution of their conservative properties.

In our experiments, we first diagnose the difference between Neural ODEs trained on very short rollouts (producing exploding trajectories in the long-term) and one trained on longer rollouts (stable), with the classical trajectory loss (eq. \ref{eq:traj_loss}), representing the gold standard. We then apply our Jacobian regularizations during training in the unstable setting (short rollouts), in order to show the improvement that our losses provide on the long-term stability of the system at inference: following \cite{white2024stabilizedneuraldifferentialequations}, we train and validate on chunks of size $N$ of short-horizon trajectories of duration $T_{\text{train}}$ and evaluate on long-term trajectories of duration $T_{\text{test}}$ (5 to 8 times longer than $T_{\text{train}}$). We compare them to a projection method SNODE \cite{white2024stabilizedneuraldifferentialequations} that achieves stable dynamics by projecting onto the manifold of algebraic constraints of the system (annex \ref{sec:snode}). Implementation details are available in annex \ref{a:implementation}.

Performing the validation on short-horizon and chunked trajectories is a particularly challenging setting for our experiments. Indeed, divergence is hard to predict when looking only at early step error, as it can happen very late during the unrolling. We chose this very restrictive protocol on purpose to be in the most realistic and reproducible scenarios possible (very expensive long-term rollouts). Finally, even though our losses require Jacobian-vector products, they avoid expensive long rollout backpropagation. In practice, training remained comparable to standard short-rollout Neural ODEs training (annex \ref{a:cost}). Additional results on the TwoBody problem experiment (annex \ref{a:tb}) confirm our results.

\begin{figure*}[t]  
  \centering
  \includegraphics[width=0.4\textwidth]{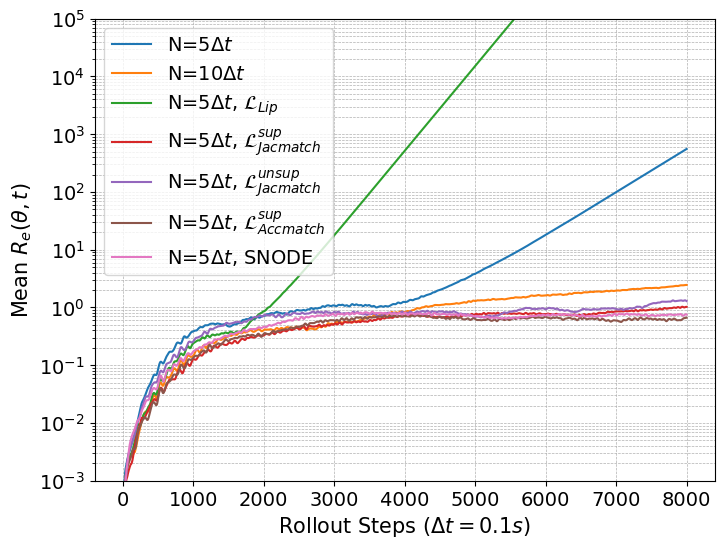}
  \includegraphics[width=0.59\textwidth]{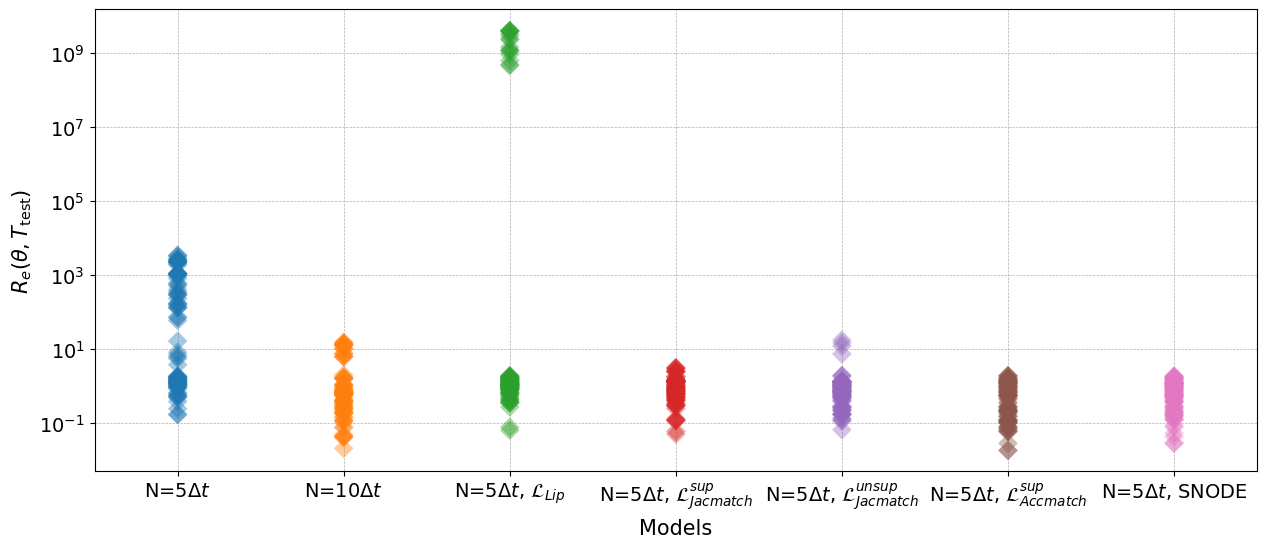}
    \includegraphics[width=0.4\textwidth]{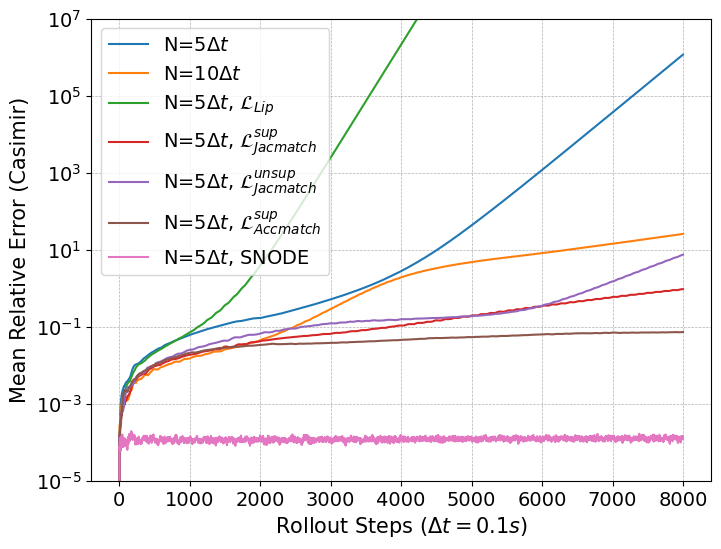}
      \includegraphics[width=0.59\textwidth]{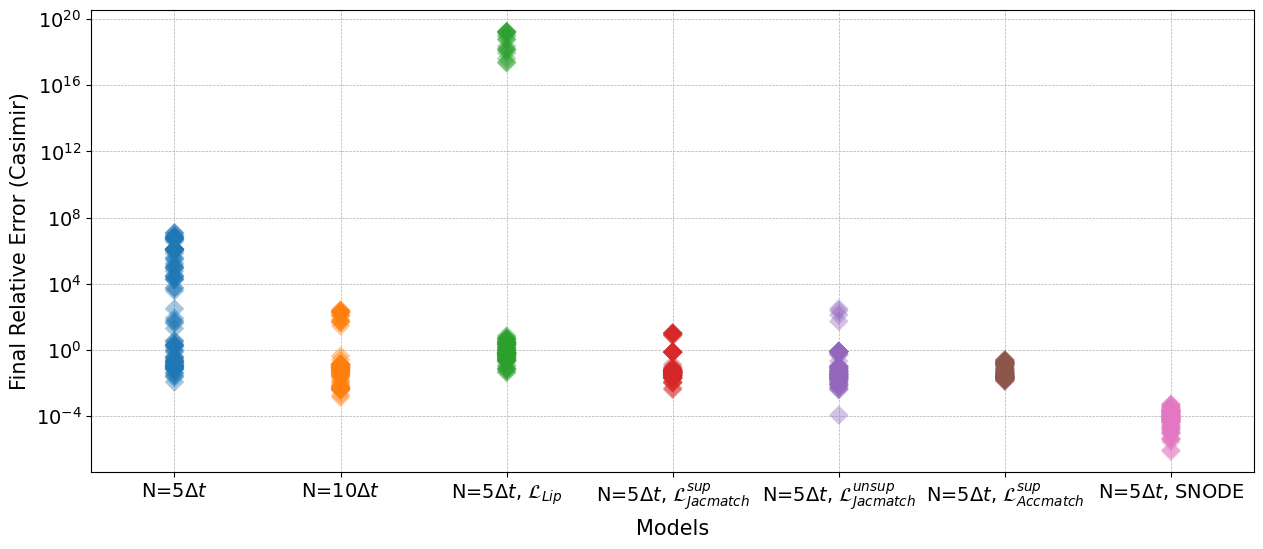}
  \caption{\textbf{Rigid Body.} For each row, we have the evolution of the relative error over time (left) and the distributions of final values (right). The top row represents the relative trajectory error $R_e(\theta,t)$ and the bottom row represents the relative conservation error (Casimir).}
  \label{fig:res-rb}
\end{figure*}

\begin{figure*}[t]  
  \centering

  \includegraphics[width=0.4\textwidth]{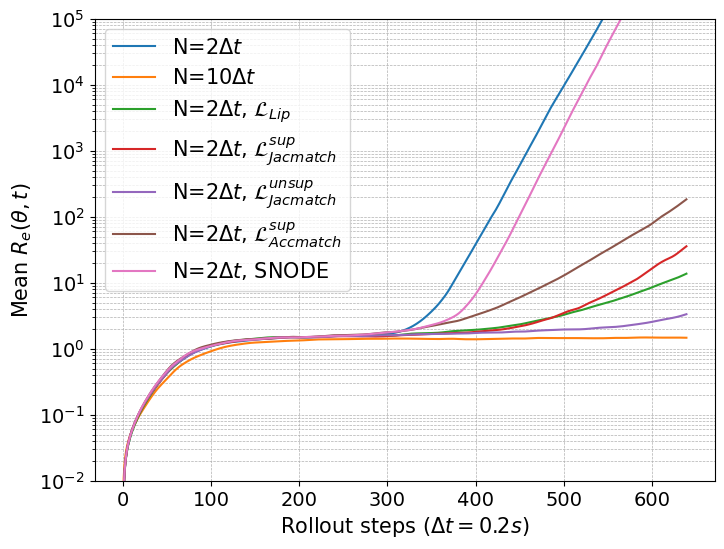}
    \includegraphics[width=0.59\textwidth]{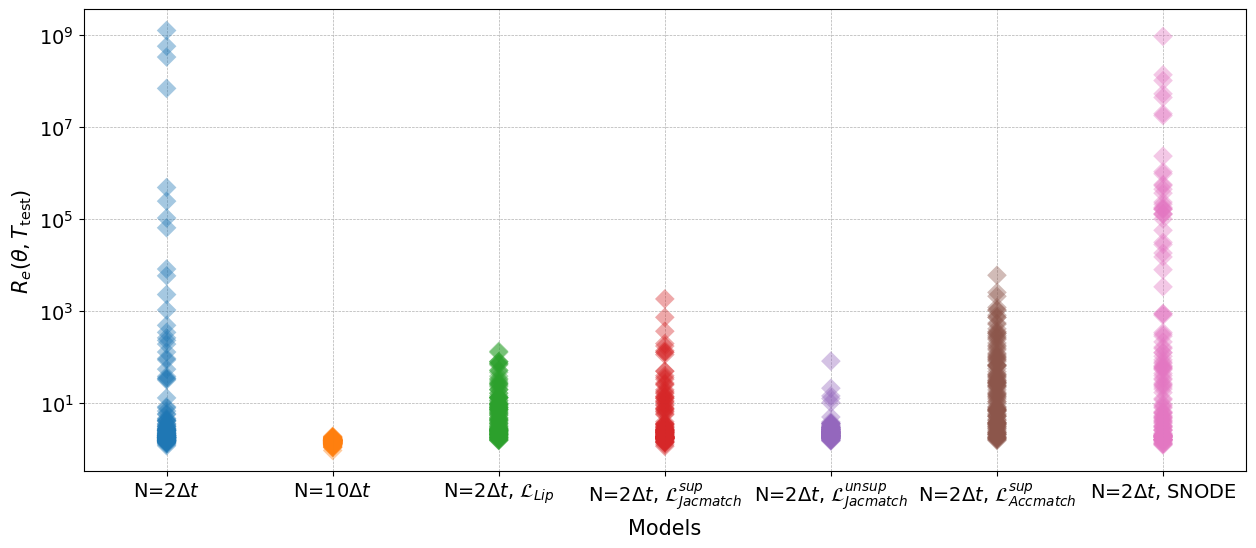}
  \includegraphics[width=0.4\textwidth]{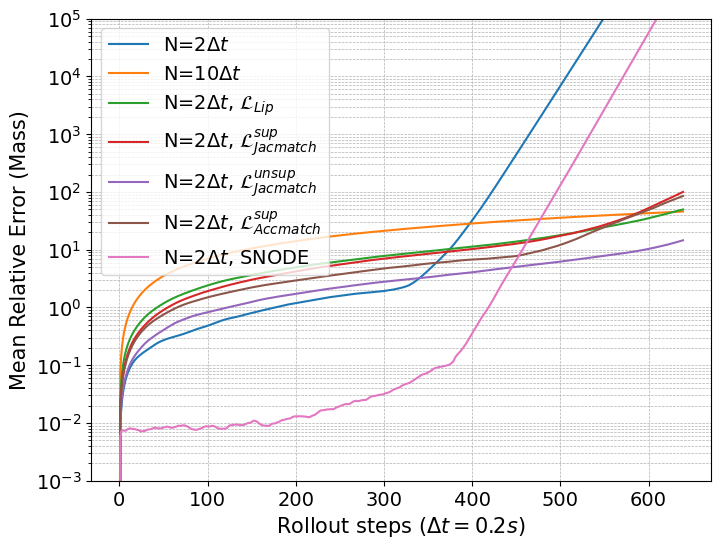}
     \includegraphics[width=0.59\textwidth]{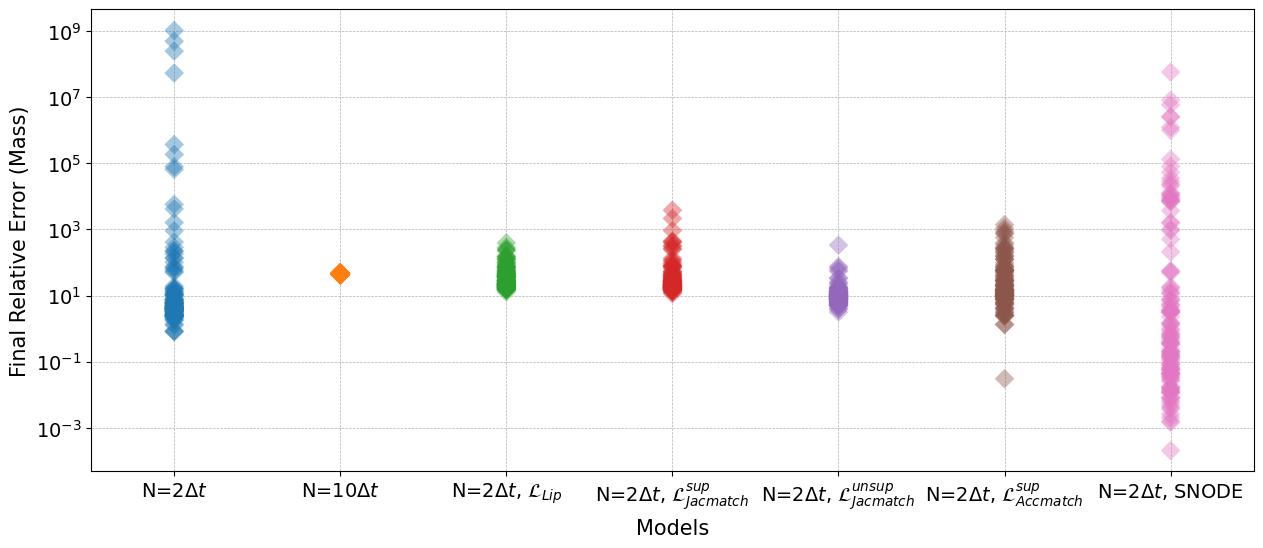}

  \caption{\textbf{Kuramoto-Sivashinsky.} For each row, we have the evolution of the relative error over time (left) and the distributions of final values (right). The top row represents the relative trajectory error $R_e(\theta,t)$ and the bottom row represents the relative conservation error (Mass).}
  \label{fig:res-ks}
\end{figure*}

As expected from \cite{frezat2021posteriorilearningquasigeostrophicturbulence, list2024differentiabilityunrolledtrainingneural}, training on longer rollouts improves long-term stability, although several trajectories still diverge after 4000 rollout steps for Rigid Body (Figure \ref{fig:res-rb}, in orange). This enhanced stability also translates into better conservation of the Casimir and Mass properties (bottom rows \ref{fig:res-rb}, \ref{fig:res-ks}). As identified in section \ref{sec:related}, it can be hard to reach an appropriate smoothness/expressiveness balance when penalizing only the norm of $J_{F_\theta}$ in $\mathcal{L}_{\text{Lip}}$ (green), especially when validating on very short rollouts as done here: this is the method that performs the worst for Rigid Body, although it is successful yet not the best for Kuramoto-Sivashinsky.

On the other hand, the $\mathcal{L}_{\text{Jacmatch}}$ regularization (extension of CFO), is successful in stabilizing the model both in supervised (red) and unsupervised (purple) settings for both physical systems. Surprisingly, the unsupervised version is even better than the supervised one for Kuramoto-Sivashinsky and even the best among all methods (Figure \ref{fig:res-ks}, left panels). This is similar to what we already observed on the impact of spline fitting in the offline experiments: the finite differences yields smoother values than the ones obtained with the true $F$ (due to the solving of spatial operations such as $u_{xx}$ or $u_{xxxx}$), facilitating training and enforcing a smoother learned representation. Our acceleration matching regularization $\mathcal{L}_{\text{Accmatch}}$ performs best in terms of trajectory error for RB, even outperforming the model trained on longer rollouts (Figure \ref{fig:res-rb}, in orange). It is also very good at conserving the Casimir property. The loss achieves relative error performances on par with the supervised $\mathcal{L}_{\text{Jacmatch}}$ and outperforms it in terms of mass conservation (Figure \ref{fig:res-ks}, bottom right) for KS. This is coherent with what we expected in section \ref{sec:method}, as the two losses are quite close in the short rollout training context (small $\epsilon_\theta(x_{t_0+N\Delta t})$ when $N$ is small).

The concurrent method, SNODE, is performing well on the small-scale system RB as in the original paper, and is the best at preserving the algebraic constraint which SNODE specifically targets. However, for KS that is a larger-scale system, the method becomes unstable: the constraint becomes too hard to maintain in the long-horizon, introducing unwanted instabilities that eventually lead to diverging trajectories. Our methods are more successful while providing a more general setup as they don't require the system to satisfy algebraic constraints. 

\section{Conclusion}
\label{sec:conclu}

In this work, we demonstrated that two seemingly distinct paradigms for stabilizing neural dynamical systems — Jacobian regularization \cite{nie2026jawsenhancinglongtermrollout} and interpolation-based training \cite{hou2025cfolearningcontinuoustimepde} — share a common theoretical foundation. By linearizing the unrolled trajectory loss, we showed that both approaches can be understood as approximations of the gradient terms that emerge from online training, specifically through their respective actions on the Jacobian norm and the Jacobian matching term. This perspective not only clarifies why these methods improve long-horizon stability, but also reveals them as complementary. Furthermore, we found that the interpolation-based CFO training leaves out the correct term that focuses on acceleration matching.

Building on this analysis, we introduced a family of losses targeting these newly identified terms, notably generalizing CFO from an interpolation strategy into a regularization. We tested these losses in both offline/short-online as well as supervised/unsupervised settings, using either exact, spline fitted or finite differences based products. All proposed methods improve on the baseline, with different degrees of success. In particular, our JacMatch approach, which generalizes CFO, is effective in stabilizing long-term dynamics in exact, spline and FD formulations. Our newly derived term AccMatch performs best offline and similarly to JacMatch in the supervised short-online setting (relevant for hybrid parametrization of physical models \cite{Kochkov_2021} and fast emulation of expensive simulators), as for short rollouts the two losses remain close. The Jacobian regularization is successful for KS although the choice of the hyperparameter $\lambda$ can remain difficult as observed for RB. 

Crucially, the unsupervised formulations of the different losses, either obtained by fitting a spline as in CFO or derived entirely from finite differences of observed trajectories, perform competitively on the different physical systems. Because it requires neither access to the true dynamics nor expensive long rollouts, it is cheap to compute and broadly applicable, suggesting potential applicability to larger-scale systems where unrolled training remains prohibitively costly. In hybrid modelling specifically, online training is often not possible due to the non-differentiability of the solvers (ocean models \cite{ouala2023onlinecalibrationdeeplearning}). 

Our theoretical analysis opens the door to defining efficient losses or protocols to stabilize the long-term dynamics learned offline on larger scale systems. The current offline framework we proposed in this paper needs to be assessed more completely by performing evaluations of the losses on other systems (for example KS, RB). Another direct perspective of our study is to apply both our offline and short-online methods on larger scale systems such as Navier-Stokes 2D \cite{yin2023continuouspdedynamicsforecasting}, quasi-geostrophic turbulence \cite{frezat2021posteriorilearningquasigeostrophicturbulence} and from other benchmarks \cite{takamoto2024pdebenchextensivebenchmarkscientific} to improve the application of our method in practice as well as our understanding of it. Finally, the methods and code to reproduce them are publicly available on GitHub, and can be directly applied to any kind of autoregressive model. 

\newpage

\section{Acknowledgments}
We would like to thank Julien Salomon (Inria, Laboratoire Jacques Louis-Lions at Sorbonne University) for his help in the various proofs of this paper. We also thank the reviewers for their insightful comments that we believe greatly improved the quality of the paper. This work is part of the PhD of Maya Janvier, funded by Sorbonne University (ED 129) and  EUR IPSL. 

\bibliography{biblio}
\bibliographystyle{plain}

\appendix
\onecolumn

\section{Notations and definitions}
\label{a:def}

\paragraph{Notations} $\Vert \cdot \Vert_2$ and $\Vert \cdot \Vert_F$ denote respectively the L2 and Fröbenius norms.

\begin{definition}
    The Jacobian of a function $F: \mathbb{R}^n \rightarrow \mathbb{R}^m$ is defined as the matrix $J_F \in \mathbb{R}^{m\times n}$ of all its first-order partial derivatives: 
\begin{equation}
        J_F 
  = \begin{pmatrix}
      \dfrac{\partial f_1}{\partial x_1} & \cdots & \dfrac{\partial f_1}{\partial x_n} \\
      \vdots                             & \ddots & \vdots                             \\
      \dfrac{\partial f_m}{\partial x_1} & \cdots & \dfrac{\partial f_m}{\partial x_n}
    \end{pmatrix}
\end{equation}

\end{definition}

\begin{definition}
A function $F$ is Lipschitz continuous if there exists $L \in \mathbb{R}^+$ such that $\forall x,y \in \mathbb{R}^n \times \mathbb{R}^n$:
\begin{equation*}
    \Vert F(x)-F(y)\Vert_2 \leq L \Vert x-y\Vert_2 .
\end{equation*}
$L$ is called a Lipschitz constant of $F$ and when it is differentiable, the optimal (smallest) Lipschitz constant is defined as $L_F = \sup_{x\in \mathbb{R}^n} \Vert J_F(x)\Vert_2$. 
\end{definition}

\section{Noise injection and Lipschitz control}
\label{a:noisy}

Let $x \in \mathbb{R}^n$ be a training point and $F: \mathbb{R}^n \rightarrow \mathbb{R}^n$ a differentiable function. 

Adversarial losses are build to resist to small perturbations or added noise (usually gaussian) $\epsilon \sim \mathcal{N}(0,I) $, that is to say to minimize $\mathcal{L}_{\text{noise}} =  \mathbb{E}_\epsilon(\Vert F(x+\epsilon)-F(x)\Vert_2$). If we use a first order Taylor-Young approximation \ref{t:taylor} in $x$, it becomes:
\begin{align*}
    \mathcal{L}_{\text{noise}} &= \mathbb{E}_\epsilon(\Vert F(x) + J_F(x)\cdot \epsilon + o(\Vert\epsilon \Vert_2)-F(x) \Vert_2 )\\
    &\leq \mathbb{E}_\epsilon( \Vert J_F(x) \cdot \epsilon \Vert_2) + \mathbb{E}_\epsilon(o(\Vert \epsilon \Vert_2)) \quad \text{(triangular inequality and linearity of expectation)} \\
    &= \Vert J_F(x) \Vert_F + \mathbb{E}_\epsilon(o(\Vert \epsilon \Vert_2)) \quad \text{(Hutchinson trace estimator)}
\end{align*}

Indeed, the Hutchinson trace estimator \cite{Hutchinson01011989} allows us to compute the Fröbenius norm of matrices using only vector products, as done in \cite{nie2026jawsenhancinglongtermrollout, Hu_2024}. Let $v \sim P(v)$ be a random vector sampled from the $P(v)$ distribution such that $\mathbb{E}_{v \sim P(v)}[vv^T]=I$ (which is the case of the normal distribution):
\begin{align}
\label{eq:hutch}
\mathbb{E}_{v \sim P(v)}[\Vert J_{F}(x) \cdot v\Vert_2^2] &= \mathbb{E}_{v \sim P(v)}[Tr(\Vert J_F(x) \cdot v \Vert_2^2)] \quad (Tr(x)=x \, \text{for $x$ scalar}) \notag  \\
&= \mathbb{E}_{v \sim P(v)}[Tr((J_{F}(x) \cdot v)^T (J_{F}(x) \cdot v))] \notag \\
&= \mathbb{E}_{v \sim P(v)}[Tr(v^T \cdot (J_{F}(x)^TJ_{F}(x)) \cdot v)]  \notag\\
&= \mathbb{E}_{v \sim P(v)}[Tr((J_{F}(x)^TJ_{F}(x)) vv^T)] \quad (Tr(AB)=Tr(BA)) \notag \\
&= Tr[ J_{F}(x)^TJ_{F}(x)\mathbb{E}_{v \sim P(v)}(vv^T)] \quad \text{(linearity expectation and trace)} \notag \\
& = Tr(J_{F}(x)^TJ_{F}(x)) \quad \text{as } \mathbb{E}_{v \sim P(v)}(vv^T)=I \notag \\
& = \Vert J_{F}(x)\Vert_F^2 \quad \text{(definition)}
\end{align}
This estimator is unbiased, and its variance is decreasing at the rate of $\frac{1}{V}$ the number of directions $v$ used when estimated through a Monte-Carlo algorithm \cite{Hu_2024}.

Furthermore, as $ \frac{1}{n} \Vert J_F(x) \Vert_F\leq \Vert J_F(x) \Vert_2 \leq \Vert J_F(x) \Vert_F$ (norms are equivalent in finite dimension), we get that using $\mathcal{L}_{\text{noise}}$ is equivalent to penalizing the Fröbenius norm of $J_F$ and ultimately the Lipschitz constant of $F$. 

\section{Lipschitz constant and trajectory error}
\label{a:lipschitz}
\textit{In this section, we denote $\Vert \cdot \Vert_2$ as $\Vert \cdot \Vert$ and $L_{F_\theta}$ as $L_\theta$ for lighter notations.}

Let us suppose we have two dynamics, the true equation with exact resolution $x$ and dynamics $F$, the other $x_\theta$ from the learned dynamic $F_\theta$:
\begin{equation}
\label{eq:basis}
\left\{
    \begin{array}{ll}
        \dot{x}(t) = F(x(t)) \\
        x(t=0) = x_0
    \end{array}
\right. \quad \text{and} \quad
\left\{
    \begin{array}{ll}
        \dot{x_\theta}(t) = F_\theta(x_\theta(t))  \\
        x_\theta(t=0) = x_0
    \end{array}
\right.
\end{equation}

We make the assumption that $F$ and $F_\theta$ are, respectively, L and $L_\theta$-Lipschitz continuous. We define $\epsilon_\theta=F-F_\theta$ the offline error function. We prove the following inequalities:

\begin{proposition}
\begin{equation}
    \forall t \geq 0, \Vert x(t)-x_\theta(t)\Vert \leq \exp(L_\theta t)  \int_0^t \Vert\epsilon_\theta(x(s))\Vert \exp(-L_\theta s)ds,
\end{equation}

and if $\epsilon_\theta$ is bounded by $\Vert\epsilon_\theta\Vert_\infty$:
\begin{equation*}
\begin{aligned}
        \Vert x(t)-x_\theta(t)\Vert &\leq \Vert\epsilon_\theta\Vert_\infty \exp(L_\theta t) \int_0^t  \exp(-L_\theta s)ds   \\
        &\leq \Vert\epsilon_\theta\Vert_\infty \exp(L_\theta t)    \frac{1}{L_\theta}(1-\exp{(-L_\theta t)})  \\
        &\leq    \frac{\Vert\epsilon_\theta\Vert_\infty}{L_\theta}(\exp{(L_\theta t)-1}).
\end{aligned}
\end{equation*}
\end{proposition}

\begin{proof}
The second inequality comes from the first one by integrating the exponential. The first one uses Grönwall's lemma:
\begin{lemma}Grönwall's inequality: \label{l:gronwall} \\
\text{Let $\psi\geq 0, \phi, K$ be continuous functions.} \\
If
$$\forall t\geq t_0, \phi(t) \leq K(t) + \int_{t_0}^t{\psi(s)\phi(s) ds},$$ 
then \\
$$\forall t\geq t_0, \phi(t) \leq K(t) + \int_{t_0}^t{K(s)\psi(s) \exp{\left( \int_s^t{\psi(u) du}  \right)} ds}.$$
\end{lemma}

As $x$ and $x_\theta$ are continuous, with the Fundamental theorem of calculus we can write: 
\begin{equation*}
\begin{aligned}
        \Vert x(t)-x_\theta(t)\Vert &= \Vert x_0 + \int_0^t{\dot{x}(s)ds} - x_0 - \int_0^t{\dot{x_\theta}(s)ds}\Vert  \\
        &= \Vert \int_0^t{F(x(s))-F_\theta(x_\theta(s)) ds} \Vert \quad \text{using eq. \ref{eq:basis}} \\
        &= \Vert \int_0^t{F(x(s))- F_\theta(x(s)) + F_\theta(x(s)) - F_\theta(x_\theta(s)) ds} \Vert \quad  \\
        &\leq  \int_0^t{\Vert\epsilon_\theta(x(s))\Vert ds}+  \int_0^t{\Vert F_\theta(x(s))-F_\theta(x_\theta(s))\Vert ds}   \quad \text{(triangular inequality)} \\
        \Vert x(t)-x_\theta(t)\Vert &\leq \int_0^t{\Vert\epsilon_\theta(x(s))\Vert ds} + L_\theta \int_0^t{\Vert x(s)-x_\theta(s)\Vert ds}  \quad \text{as $F_\theta$ is $L_\theta$-Lipschitz.}
\end{aligned}
\end{equation*}

We can apply Grönwall's inequality with $\phi(t) = \Vert x(t)-x_\theta(t)\Vert$, $\psi(t)= L_\theta \geq 0$ and $K(t)= \int_0^t{\Vert\epsilon_\theta(x(s))\Vert ds}$:

\begin{equation*}
\begin{aligned}
        \Vert x(t)-x_\theta(t)\Vert
        &\leq \int_0^t{\Vert\epsilon_\theta(x(s))\Vert ds} + \int_0^t{ L_\theta\ \left( \int_0^s{\Vert\epsilon_\theta(x(u))\Vert du}\right) \exp{\left( \int_s^t{L_\theta du}  \right) ds} }\\
        &\leq K(t) + L_\theta \exp{(L_\theta t)}\int_0^t{ \left( \int_0^s{\Vert\epsilon_\theta(x(u)\Vert du}\right) \exp{(-L_\theta s)}ds} .
\end{aligned}
\end{equation*}

For the integral, we can use integration by parts (IPP), with

\begin{equation*}
\left\{
    \begin{array}{ll}
        h(s)= \int_0^s{\Vert\epsilon_\theta(x(u))\Vert du} =K(s)\\
        \dot{h}(s) = \Vert\epsilon_\theta(x(s))\Vert
    \end{array}
\right. \quad \text{and} \quad
\left\{
    \begin{array}{ll}
        g(s) = -\frac{1}{L_\theta}\exp(-L_\theta s) \\
        \dot{g}(s)=\exp{(-L_\theta s)} 
    \end{array}
\right.
\end{equation*}

We have: 
\begin{equation*}
\begin{aligned}
       \int_0^t{h(s)\dot{g}(s)ds} &= \left[ h(s)g(s) \right]_0^t - \int_0^t{\dot{h(s)}g(s) ds} \\ 
       &= \left[ K(s) \frac{-1}{L_\theta}\exp(-L_\theta s) \right]_0^t - \int_0^t \frac{-1}{L_\theta}\exp(-L_\theta s)\Vert\epsilon_\theta(x(s))\Vert ds \\
       \int_0^t{h(s)\dot{g}(s)ds}&= \frac{-1}{L_\theta}\exp(-L_\theta t) K(t) + \frac{1}{L_\theta}\int_0^t \Vert\epsilon_\theta(x(s))\Vert\exp(-L_\theta s)ds
\end{aligned}
\end{equation*}

Plugging it back to the main inequality,
\begin{equation*}
\begin{aligned}
        \Vert x(t)-x_\theta(t)\Vert &\leq K(t) + L_\theta \exp{(L_\theta t)} \left( \int_0^t{\left(\int_0^s{\Vert\epsilon_\theta(x(u))\Vert du}\right)  \exp{(-L_\theta s)}ds} \right) \\
        &\leq K(t) + L_\theta \exp(L_\theta t) \left( \frac{-1}{L_\theta}\exp(-L_\theta t)K(t) + \frac{1}{L_\theta} \int_0^t \Vert\epsilon_\theta(x(s))\Vert \exp(-L_\theta s)ds\right) \\
        &\leq K(t)  -K(t) + \exp(L_\theta t)\int_0^t \Vert\epsilon_\theta(x(s))\Vert \exp(-L_\theta s)ds \\
        &\leq  \exp(L_\theta t)  \int_0^t \Vert\epsilon_\theta(x(s))\Vert \exp(-L_\theta s)ds.
\end{aligned}
\end{equation*}
\end{proof}

\section{Effect of unrolling}
\label{a:unrolling}

We consider the same notations and definitions that in annex \ref{a:lipschitz}. We suppose that $F$ and $F_\theta$ are differentiable, in order to use the Taylor-Young formula:
\begin{theorem} Taylor-Young formula
\label{t:taylor}

If $f:\mathbb{R}^n \rightarrow \mathbb{R}^m$ is differentiable $k$ times in $a\in\mathbb{R}^n$, we have
\begin{equation}
    f(x) = \sum_{j=0}^{k}
           \frac{f^{(j)}(a)}{j!}(x-a)^j
           + o\left((x-a)^k\right), \quad x \to a
\end{equation}  
\end{theorem}

\subsection{Euler forward $x_t \neq x_t^\theta$}

Let's derive equation \ref{eq:unrolling}. Starting from $x_t$, if the solver is Euler forward we have:
\begin{align*}
x_{t+2\Delta t} &= x_{t+\Delta t} + F(x_{t+\Delta t})\,\Delta t \,  \\
x_{t+2\Delta t}    &= x_t + F(x_t)\,\Delta t + F\!\left(x_t + F(x_t)\,\Delta t \right)\Delta t \, 
\end{align*}
Similarly, starting from some $x_t^\theta$ with the same solver:
\begin{equation*}
x^\theta_{t+2\Delta t} = x^\theta_t + F_\theta(x^\theta_t)\,\Delta t + F_\theta\!\left(x^\theta_t + F_\theta(x^{\theta}_t)\,\Delta t \right)\Delta t \, 
\end{equation*}

Thus, the error between the two is: 
\begin{align}
\label{eq:error1}
 x_{t+2\Delta t} - x^\theta_{t+2\Delta t}
&= x_t - x^\theta_t
+ \Delta t\!\left(F(x_t) - F_\theta(x^\theta_t)\right) \notag \\
&+ \Delta t\!\left(F\!\left(x_t + F(x_t)\,\Delta t\right) - F_\theta\!\left(x^\theta_t + F_\theta(x^\theta_t)\,\Delta t\right)\right) \notag 
\end{align}

Using Taylor-Young formula we get:
\begin{align*}
F(x_t) - F_\theta(x^\theta_t)
&= F(x_t) - F_\theta(x_t) + J_{F_\theta}(x_t)\cdot(x_t - x^\theta_t) \, + o(\Vert x_t-x_t^\theta \Vert_2) \\
F\!\left(x_t + F(x_t)\,\Delta t\right)
&= F(x_t) + J_F(x_t)\cdot F(x_t)\,\Delta t + o(\Delta t)\\
F_\theta\!\left(x^\theta_t + \,F_\theta(x^\theta_t)\Delta t\right) 
&= F_\theta(x_t) + J_{F_\theta}(x_t)\cdot\left(x^\theta_t + F_\theta(x^\theta_t)\,\Delta t - x_t\right) +o(\Delta t) + o(\Vert x_t-x_t^\theta \Vert_2) \\
&= F_\theta(x_t) + J_{F_\theta}(x_t)\cdot(x^\theta_t - x_t) + J_{F_\theta}(x_t)\cdot F_\theta(x^\theta_t)\,\Delta t + o(\Delta t) + o(\Vert x_t-x_t^\theta \Vert_2)
\end{align*}

Substituting back into equation \ref{eq:error1} :
\begin{align*}
x_{t+2\Delta t} - x^\theta_{t+2\Delta t}
&= x_t - x^\theta_t
+ \Delta t\!\left(\epsilon_\theta(x_t) + J_{F_\theta}(x_t)\cdot(x_t - x^\theta_t) + o(\Vert x_t-x_t^\theta \Vert_2)\right) \notag \\
&\quad + \Delta t\!\Bigl(F(x_t) + J_F(x_t)\cdot F(x_t)\,\Delta t
  - F_\theta(x_t) - J_{F_\theta}(x_t)\cdot(x^\theta_t - x_t)
  - J_{F_\theta}(x_t)\cdot F_\theta(x^\theta_t)\,\Delta t \notag \\
  &\quad + o(\Delta t) + o(\Vert x_t-x_t^\theta \Vert_2) \Bigr) \notag\\
&= x_t - x^\theta_t
+ 2\,\Delta t\, \big(\epsilon_\theta(x_t)
+  J_{F_\theta}(x_t)\cdot(x_t - x^\theta_t) + o(\Vert x_t-x_t^\theta \Vert_2) \big) \\
&\quad +\Delta t^2\!\left(J_F(x_t)\cdot F(x_t) - J_{F_\theta}(x_t)\cdot F_\theta(x^\theta_t) + o(\Vert x_t-x_t^\theta \Vert_2)\right) \\
& \quad + o(\Delta t^2) 
\end{align*}

Now we expand the $\Delta t^2$ term:
\begin{align}
&J_F(x_t)\cdot F(x_t) - J_{F_\theta}(x_t)\cdot F(x_t) + J_{F_\theta}(x_t)\cdot F(x_t) - J_{F_\theta}(x_t)\cdot F_\theta(x^\theta_t) \notag\\
&= J_{F - F_\theta}(x_t)\cdot F(x_t) + J_{F_\theta}(x_t)\cdot\!\left(F(x_t) - F_\theta(x^\theta_t)\right) \notag\\
&= J_{\epsilon_\theta}(x_t)\cdot F(x_t) + J_{F_\theta}(x_t)\cdot\!\left(F(x_t) - F_\theta(x_t) + F_\theta(x_t) - F_\theta(x^\theta_t)\right) \notag\\
&= J_{\epsilon_\theta}(x_t)\cdot F(x_t)
+ J_{F_\theta}(x_t)\cdot \epsilon_\theta(x_t)
+ J_{F_\theta}(x_t)\cdot\!\left(F_\theta(x_t) - F_\theta(x^\theta_t)\right)
\end{align}

Substituting everything back, we finally get:
\begin{align}
x_{t+2\Delta t} - x^\theta_{t+2\Delta t}
&= (x_t - x^\theta_t) \notag\\
&\quad + 2\,\Delta t\, \big( \epsilon_\theta(x_t) + J_{F_\theta}(x_t)\cdot(x_t - x^\theta_t) + o(\Vert x_t-x_t^\theta\Vert_2) \big) \notag\\
&\quad + \Delta t^2\, \big[ J_{\epsilon_\theta}(x_t)\cdot F(x_t) + J_{F_\theta}(x_t)\cdot \epsilon_\theta(x_t)  \notag\\
&\quad + \,J_{F_\theta}(x_t)\cdot\bigl(F_\theta(x_t) - F_\theta(x^\theta_t)\bigr) + o(\Vert x_t-x_t^\theta \Vert_2) \big]\notag \\
&\quad + o(\Delta t^2) 
\end{align}

\subsection{Euler forward $x_t=x_t^\theta$}
Now if $x_t=x_t^\theta$ (in the training context), we have:
\begin{align}
x_{t+2\Delta t} - x^\theta_{t+2\Delta t}
&=  2\,\Delta t\, \epsilon_\theta(x_t) + \Delta t^2\, \big( J_F(x_t)\cdot F(x_t) -J_{F_\theta}(x_t)\cdot F_\theta(x_t) \big) + o(\Delta t^2) \\
&= 2\,\Delta t\, \epsilon_\theta(x_t) + \Delta t^2\, \eta_\theta(x_t) + o(\Delta t^2)
\end{align}

We can generalize the formula to $N$ steps by induction. We define:  
\begin{align}
    H_N: \left\{ \begin{array}{ll}
         x_{t+N\Delta t} &= x_t + N\Delta t F(x_t) + \frac{N(N-1)}{2} J_F(x_t)\cdot F(x_t) + o(\Delta t^2) \\
         x_{t+N\Delta t}^\theta &= x_t + N\Delta t F_\theta(x_t) + \frac{N(N-1)}{2} J_{F_\theta}(x_t)\cdot F_\theta(x_t) + o(\Delta t^2)
    \end{array}
    \right. \\
    \text{such that  }x_{t+N\Delta t} - x_{t+N\Delta t}^\theta =N\Delta t \epsilon_\theta(x_t) + \frac{N(N-1)}{2} \eta_\theta(x_t) + o(\Delta t^2), \quad n\geq 2
\end{align}

The formula is already true for $N=2$, we suppose $H_N$ is true for a certain $N\geq 2$. We have under Euler forward:
\begin{align*}
    x_{t+(N+1)\Delta t} -x_{t+(N+1)\Delta t}^\theta=x_{t+N\Delta t} - x_{t+N\Delta t}^\theta + \Delta t ( F(x_{t+N\Delta t}) - F_\theta(x_{t+N\Delta t}^\theta))
\end{align*}
and under $H_N$:
\begin{align*}
    x_{t+N\Delta t} &= x_t + F(x_t)\Delta t + F(x_t+F(x_t)\Delta t)\Delta t + ... \\
    &= x_t + \delta_N \\
    \delta_N &= N\Delta t F(x_t) + \frac{N(N-1)}{2}  \Delta t^2 J_F(x_t) \cdot F(x_t) + o(\Delta t^2) \\
    &= N\Delta t F(x_t) +o(\Delta t) 
\end{align*}
By using Taylor's formula around $x_t$:
\begin{align*}
    F(x_{t+N\Delta t}) &= F(x_t+\delta_N) \\
    &= F(x_t) + J_F(x_t)\cdot \delta_N + o(\Vert \delta_N\Vert_2^2) \\
    &= F(x_t) + J_F(x_t)\cdot (N\Delta F(x_t)+ o(\Delta t)) + o(\Delta t^2) \\
    & = F(x_t) + N\Delta t J_F(x_t)\cdot F(x_t) + o(\Delta t) 
\end{align*}
and 
\begin{align*}
    x_{t+(N+1)\Delta t }&= x_t + (N+1)\Delta tF(x_t) + (\frac{N(N-1)}{2}+ N)J_F(x_t) \cdot F(x_t) + o(\Delta t^2) \\
    &=x_t + (N+1)\Delta tF(x_t) + \frac{N(N+1)}{2} J_F(x_t) \cdot F(x_t) + o(\Delta t^2)   
\end{align*}

The coefficients are the same for $x_{t+(N+1)\Delta t}^\theta=x_t + F_\theta(x_t)\Delta t + F_\theta(x_t+F_\theta(x_t)\Delta t)\Delta t + ...$, only with $F_\theta$ since $x_t=x^\theta_t$, and we get $H_{N+1}$:
\begin{align}
    x_{t+(N+1)\Delta t} - x_{t+(N+1)\Delta t}^\theta = (N+1) \epsilon_\theta(x_t) + \frac{N(N+1)}{2} \eta_\theta(x_t) + o(\Delta t^2)
\end{align}

\subsection{RK2 and higher-order explicit schemes}
\label{a:rk2}
The precedent reasoning can be applied to other integration schemes. For RK2, we have 
\begin{align*}
    x_{t+\Delta t} &= x_t + \Delta t F(x_t+ \frac{\Delta t}{2}F(x_t)) \\
    &= x_t + \Delta t (F(x_t) + J_F(x_t) \cdot \frac{\Delta t}{2}F(x_t) + o(\Delta t)) \\ 
    &= x_t + \Delta t F(x_t) + \frac{\Delta t^2}{2} J_F(x_t) \cdot F(x_t) + o(\Delta t^2)
\end{align*}
and 
\begin{equation}
    \label{eq:rk2}
    x_{t+\Delta t}-x_{t+\Delta t}^\theta = \Delta t \epsilon_\theta(x_t) + \frac{\Delta t^2}{2} \eta_\theta(x_t) + o(\Delta t^2)
\end{equation}
A similar induction than the previous section will give the general formula:
\begin{equation}
    x_{t+N\Delta t}-x_{t+ N\Delta t}^\theta = N\Delta t \epsilon_\theta(x_t) + \frac{N^2\Delta t^2}{2} \eta_\theta(x_t) + o(\Delta t^2)
\end{equation}
For higher order Runge-Kutta methods, if we limit ourselves to $o(\Delta t^2)$ order terms, RK4 will give the same terms as RK2 (but differ when we look at higher order terms). We observe that when using RK2, eq. \ref{eq:rk2} recovers the formula we would get from applying Taylor-Young to $x(t+\Delta t)$ in $t$ at the order 2. 

\subsection{Unrolled gradients for RK2}
\label{a:unroll_grad}
We can now analyze the regularity needed for $F, F_\theta$ and their derivatives in order for our approximation to be close to the online training loss. 

We denote $r_\theta=\Delta t \epsilon_\theta(x_t) + \frac{\Delta t^2}{2} \eta_\theta(x_t)$ so that we derive from eq. \ref{eq:rk2} the approximated loss $\mathcal{L}_{\text{approx}}(\theta)= \Vert r_\theta\Vert_2^2$, that aims at approaching $\mathcal{L}^{N=1}_{\text{traj}}(\theta)=\Vert x_{t+\Delta t}-x_{t+\Delta t}^\theta \Vert_2^2$ in the case where $x_t^\theta=x_t$. From eq. \ref{eq:rk2}, we define $\delta_\theta =(x_{t+\Delta t}-x_{t+\Delta t}^\theta) - r_\theta$, so we have $\mathcal{L}^{N=1}_{\text{traj}}(\theta)=\Vert r_\theta+\delta_\theta \Vert_2^2$. With our previous assumptions ($F, F_\theta$ differentiable), we know that $\delta_\theta=o(\Delta t^2)$.

\paragraph{Assumptions} For $\theta$ in a compact set $\Theta$, and $K$ a compact containing $x_t$ and the trajectories starting from it when $F_\theta$ spans $\Theta$, we suppose that:
\begin{itemize}
    \item (H1) $F$ is $C^2$ in $x$.
    \item (H2) $F_\theta$ is $C^3$ in $x$, and $D^k_xF_\theta$ ($k\le 2$) is $C^1$ in $\theta$ with jointly
    continuous derivatives. Hence $D^3F_\theta$, $\partial_\theta F_\theta$, $\partial_\theta DF_\theta$,
    $\partial_\theta D^2F_\theta$ are continuous, thus bounded on $K\times\Theta$.
\end{itemize}
where the $D, D^2$ notations correspond to first, second order derivatives. 

\paragraph{Characterization of $\delta_\theta$} Since $F$ is $C^2$, $x$ is $C^3$ and we have with Taylor-Young: 
\begin{equation}
    x_{t+\Delta t} = x_t + \Delta t F(x_t)+\frac{\Delta t^2}{2} J_F(x_t)\cdot F(x_t) + r_F, \quad r_F=\frac{\Delta t^3}{6} x'''(t) + o(\Delta t^3) = O(\Delta t^3).
\end{equation}

For $x^\theta_{t+\Delta t}$, since we will need to differentiate $\delta_\theta$ through $\theta$, we need to write the RK2 update using Taylor's theorem with first order integral remainder: 
\begin{align}
    x_{t+\Delta t}^\theta &= x_t + \Delta t F_\theta(x_t + \frac{\Delta t}{2}F_\theta(x_t)) \notag \\ 
    &= x_t +\Delta t (F_\theta(x_t) + \frac{\Delta t}{2} J_{F_\theta}(x_t) \cdot F_\theta(x_t) + r_{1}^\theta) \notag \\
    &= x_t + \Delta t F_\theta(x_t) + \frac{\Delta t^2}{2} J_{F_\theta}(x_t) \cdot F_\theta(x_t) + \Delta t r_1^\theta 
\end{align}
with $r_1^\theta$ the first order integral remainder. Writing $h_\theta=\frac{\Delta t}{2}F_\theta(x_t)$, it is defined as:
\begin{align}
    r_1^\theta &=\int_0^1(1-s)D^2F_\theta(x_t+ sh_\theta)[h_\theta, h_\theta] ds \notag \\ 
    \label{eq:remainder}
    &= \frac{\Delta t^2}{4} \int_0^1 (1-s) D^2F_\theta(x_t+ sh_\theta)[F_\theta(x_t), F_\theta(x_t)] ds.
\end{align}
We then get a formula for $\delta_\theta=r_F-\Delta tr_1^\theta$. 

\paragraph{Order of the gradient approximation} We derive the gradient of $\mathcal{L}_\text{traj}^{N=1}$:
\begin{align}
    \frac{\partial \mathcal{L}_\text{traj}^{N=1}(\theta)}{\partial \theta} &= \frac{\partial \Vert r_\theta + \delta_\theta \Vert_2^2}{\partial \theta }  \notag \\ 
    &= 2(r_\theta+\delta_\theta)^T \frac{\partial}{\partial \theta}(r_\theta+\delta_\theta) \notag \\
    &=2r_\theta^T\frac{\partial r_\theta}{\partial \theta} + 2\delta_\theta^T\frac{\partial r_\theta}{\partial \theta} + 2r_\theta^T\frac{\partial \delta_\theta}{\partial \theta} + 2\delta_\theta^T\frac{\partial \delta_\theta}{\partial \theta} \notag \\
    &= \frac{\partial \mathcal{L}_\text{approx}}{\partial \theta }+ 2\delta_\theta^T\frac{\partial r_\theta}{\partial \theta} + 2r_\theta^T\frac{\partial \delta_\theta}{\partial \theta} + 2\delta_\theta^T\frac{\partial \delta_\theta}{\partial \theta}
.\end{align}
Let us bound the different terms: 
\begin{itemize}
    \item $r_\theta=\Delta t (\epsilon_\theta(x_t) + \frac{\Delta t}{2} \eta_\theta(x_t))$ is clearly $O(\Delta t)$;
    
    \item $\dfrac{\partial r_\theta}{\partial\theta}=-\Delta t\Big(\partial_\theta F_\theta+\dfrac{\Delta t}{2}\,
    \partial_\theta\big(J_{F_\theta}F_\theta\big)\Big)$, with
    $\partial_\theta(J_{F_\theta}F_\theta)=(\partial_\theta J_{F_\theta})F_\theta+J_{F_\theta}\partial_\theta F_\theta$,
    all evaluated at $x_t$. By assumption (H2) this is $O(\Delta t)$;
    
    \item $\Vert r_1^\theta \Vert \leq \frac{\Delta t^2}{8} \sup_{K} \Vert D^2 F_\theta \Vert \Vert F_\theta(x_t)\Vert^2$, so $r_1^\theta=O(\Delta t^2)$, and since $r_F=O(\Delta t^3)$, we have that $\delta_\theta = r_F-\Delta t r_1^\theta=O(\Delta t^3)$;
    
    \item $r_F$ does not depend on $\theta$, so $\dfrac{\partial\delta_\theta}{\partial\theta}=-\Delta t\,\partial_\theta r_1^\theta$.
    Under (H2) we can use Leibniz rule to differentiate under the integral in eq. \ref{eq:remainder}, and using the symmetry of $D^2F_\theta$, we get
    \begin{align}
        \partial_\theta r_1^\theta=\frac{\Delta t^2}{4}\int_0^1(1-s)\Big\{
        &(\partial_\theta D^2F_\theta)(x_t+sh_\theta)[F_\theta,F_\theta]
        +s\,\frac{\Delta t}{2}\,D^3F_\theta(x_t+sh_\theta)[\partial_\theta F_\theta,F_\theta,F_\theta] \notag\\
        &+2\,D^2F_\theta(x_t+sh_\theta)[\partial_\theta F_\theta,F_\theta]\Big\}\,ds,
    \end{align}
    where $F_\theta,\partial_\theta F_\theta$ are evaluated at $x_t$. The three terms under the integral are bounded under (H2), and are respectively $O(\Delta t^2)$, $O(\Delta t^3)$, $O(\Delta t^2)$. Hence
    $\partial_\theta r_1^\theta=O(\Delta t^2)$ and $\dfrac{\partial\delta_\theta}{\partial\theta}=O(\Delta t^3)$.
\end{itemize}

\paragraph{Conclusion} Combining the bounds,
\begin{equation}
    \Big\|2\delta_\theta^T\frac{\partial r_\theta}{\partial\theta}\Big\|=O(\Delta t^4),
    \Big\|2r_\theta^T\frac{\partial\delta_\theta}{\partial\theta}\Big\|=O(\Delta t^4), 
    \Big\|2\delta_\theta^T\frac{\partial\delta_\theta}{\partial\theta}\Big\|=O(\Delta t^6). \notag
\end{equation}
Therefore,
\begin{equation}
    \Big\|\frac{\partial\mathcal{L}^{N=1}_{\text{traj}}}{\partial\theta}-\frac{\partial\mathcal{L}_{\text{approx}}}{\partial\theta}\Big\|\le C\,\Delta t^4 .
\end{equation}
Since $\frac{\partial\mathcal{L}_{\text{approx}}}{\partial\theta}=2r_\theta^T\frac{\partial r_\theta}{\partial\theta}=O(\Delta t^2)$,
the gradient of $\mathcal{L}_{\text{approx}}$ approximates that of $\mathcal{L}^{N=1}_{\text{traj}}$
with relative error $O(\Delta t^2)$.

\section{More on the competing methods}
\label{a:competing}

\subsection{SNODE \cite{white2024stabilizedneuraldifferentialequations}}
\label{sec:snode}
The idea of SNODE is to directly modify the Neural ODE $\frac{du}{dt}=f_\theta(u)$ with a term that “nudges” back the trajectory to the manifold of algebraic constraint $\mathcal{M}=\{u/ g(u)=0\}$: 
\begin{equation}
\label{eq:snode}
    \frac{du}{dt}=f_\theta(u) - \gamma F(u)g(u)
\end{equation}
Equation \ref{eq:snode} is called the Stabilized NODE. They prove that the manifold $\mathcal{M}$ is asymptotically Lyapunov-stable in the SNODE under certain constraints for the stabilization matrix $F$, which they choose to be the Moore-Penrose pseudoinverse of the Jacobian of $g$ at $u$. Pal et al. (2025) \cite{pal2025semiexplicitneuraldaeslearning} showed that this setting was equivalent to only doing the first iteration of the projection (solved as a fixed point problem) onto the manifold. 

This approach is limited to systems that respect algebraic constraints that are verified for each timestep (ex: a system under a temporal forcing). Secondly, when the system is chaotic like KS, it becomes hard to find a value for $\gamma$ that allows the stability criteria to be valid throughout long rollouts (see eq.10 in the original paper and our experiment in this paper). Another limitation comes from the computational cost of the pseudoinverse in $O(m^2n)$ (where $m$ is the number of constraints and $n$ is the dimension of the problem) which becomes prohibitively expensive to compute in higher dimensions, whereas our methods only require evaluations of $F_\theta$ and $F$.

\subsection{JAWS \cite{nie2026jawsenhancinglongtermrollout}}
\label{sec:jaws}
JAWS is cited in this paper as a representative of the Jacobian regularization methods, that typically involve the core $\mathcal{L}_\text{Lip} = \Vert J_{F_\theta}(x_t)\Vert_2^2$ term obtained via the Hutchinson estimator. The JAWS loss is : 
\begin{equation}
    \mathcal{L} = \sum_{x \in \Omega} \frac{1}{2} e^{-s1} \Vert x_{t+1}-x_{t+1}^\theta \Vert^2  + \frac{1}{2} e^{-s2} \Vert J_{F_\theta}(x)\Vert^2_F + \frac{1}{2} (s_1+s_2) +\lambda \sum_{k=2}^K \Vert x_{t+k}-x_{t+k}^\theta \Vert^2
\end{equation}

It combines the trajectory loss (term weighted by $\lambda$) and an adaptative reconstruction/Jacobian regularization loss with balanced $s_1,s_2$ weights. The method is equivalent to $\mathcal{L}_\text{Lip}$ except for the adaptative coefficient strategy. 

\subsection{CFO \cite{hou2025cfolearningcontinuoustimepde}}
\label{sec:cfo}

CFO is a fully offline training method that belongs to interpolation based/data augmentation strategies, and relies on 3 concepts: 
\begin{itemize}
    \item building a spline approximation of $F$ (done once) based on FD of trajectory points
    \item super-sample (data augmentation) using this spline during training: introduces the JacMatch regularization as demonstrated in eq. \ref{eq:cfo-lin}
    \item noise scheduler: the points are sampled along a noisy path (path = spline + $\gamma(t) \epsilon$ with $\epsilon \approx N(0,I)$). This penalises $||J_{\epsilon_\theta}||_F$ (Hutchinson) thus bringing the Lipschitz constants of $F, F_\theta$ closer (annex \ref{a:noisy})
\end{itemize}

Its loss is: 
\begin{equation}
    \mathcal{L}_{\text{CFO}}(\theta)=\sum_{X\in \mathcal{T}_{\text{I}}} \Vert F_\theta(x(t_i))-\partial_tI(t_i,x(t_i))\Vert^2_2
\end{equation}
where $I$ is the fitted spline and $\mathcal{T}_{\text{I}}$ the ensemble of data points sampled along $I$. The novelty in CFO lies mostly in this fitting+super-sampling that we linked to JacMatch in \ref{eq:cfo-lin}. The cost of the method can grow with the number of points used for the upsampling. 

\section{Implementation details}
\label{a:implementation}
This section presents the different parameters and protocols to reproduce our experiments. The code is implemented in jax and available on GitHub. 

\subsection{Data generation} 

\paragraph{Van der Pol} The Van der Pol (VdP) oscillator is a non-conservative, oscillating system with non-linear damping. It evolves in time according to the second order differential equation:

\begin{equation}
    \frac{dx^2}{dt^2}-\mu(1-x^2)\frac{dx}{dt}+x=0
\end{equation}
with $\mu=1$. The initial conditions $x_0=(u_0,v_0)$ are randomly sampled from the uniform distribution between $[-2.5, 2.5]^2$, and the ground truth trajectories simulated with $\Delta t=0.01s$. The majority of the results are reported for a model timestep of $0.1s$, 10 different trajectories for each train/val/test sets and $T_{\text{train}}=T_{\text{test}}=10s$.

\paragraph{Rigid Body} The Rigid Body (RB) equations describe the evolution of the angular momentum vector $X$ of a rigid body under Euler's motion laws \cite{hairer_geometricnumericalintegration}: 
\begin{equation}
    U = \begin{pmatrix}
    y_1 \\
    y_2 \\
    y_3
\end{pmatrix}, \dot{U} = \begin{pmatrix}
    0 & -y_3 & y_2 \\
    y_3 & 0 & -y_1 \\
    -y_2 & y_1 & 0 
\end{pmatrix} 
\begin{pmatrix}
    y_1/I_1 \\
    y_2/I_2 \\
    y_3/I_3
\end{pmatrix}
\end{equation}
with $I_1, I_2, I_3$ the principal moments of inertia, where the coordinate axes are the principal axes of the body and the origin of the coordinates fixed at the center of mass of the body \cite{white2024stabilizedneuraldifferentialequations}. The Casimir function $C(U)=\frac{1}{2}(y_1^2+y_2^2+y_3^2$) is conserved along the trajectory and constitutes a holonomic constraint.
Similarly to White \cite{white2024stabilizedneuraldifferentialequations}, we train on 40 trajectories with initial conditions $X_0=(\cos{\phi}, 0, \sin{\phi})$ with $\phi$ sampled from $ \mathcal{U}(0.5,1.5)$. We use data from trajectories of $T_{\text{train}}=15s$ integrated from $F_{\text{RB}}$ with a time-step $\Delta t=0.01s$. We then divide these trajectories into non-overlapping chunks of $N=5\Delta t$ length. The ground truth and the models both use RK4 but with different time-steps: $\Delta t_{\text{model}}=10\Delta t=0.1s$ for faster training. We then evaluate the long-term stability of our models on 100 test trajectories of duration $T_{\text{test}}=800s$. We use the same parameter values $(I_1, I_2, I_3)=(1.6, 1.0, 2/3)$ as in White \cite{white2024stabilizedneuraldifferentialequations}.

\paragraph{Kuramoto-Sivashinksy} The 1D Kuramoto-Sivashinsky (KS) equation is a fourth order chaotic PDE describing the instabilities of a laminar flame front:
\begin{equation}
    U_t + U_{xx}+ U_{xxxx}+ UU_x=0
\end{equation}
Apart from being non linear, the biharmonic term is particularly difficult to model. Along a trajectory, we have $\int_0^LU(x,t)=0$, $L$ being the length of the spatial domain with periodic boundary conditions. To generate ground truth data, we follow the setups of \cite{brandstetter2022liepointsymmetrydata} and \cite{lippe2023pderefinerachievingaccuratelong}, using a uniform discretization of the domain into 256 points. We do not sample across different $L$ and $\Delta t$ but keep them fixed at $L=64, \Delta t=0.2$, and generate 512 training trajectories of duration $T_{\text{train}}=140\Delta t$. We alter the generation script of \cite{brandstetter2022liepointsymmetrydata}  by fixing $\Delta t=0.2s$ and $L=64$, instead of varying them. The code can then be used to produce our data with the following arguments: 
\begin{itemize}
    \item Training set: \texttt{--train\_samples=512 --nt=500}
    \item Validation set: \texttt{--valid\_samples=128 --nt=500}
    \item Test set:  \texttt{--test\_samples=128 --nt=1000 --nt\_effective=640 --end\_time=200}
\end{itemize}
As in \cite{lippe2023pderefinerachievingaccuratelong}, the first 360 steps are truncated and considered as a warmup for the solver (i.e., $t_0=360\Delta t$). The initial conditions are sampled from a distribution over truncated Fourier series with random
coefficients $\{A_m,l_m,\phi_m\}$ as $U_0(x) = \sum_{m=1}^{10} A_m \sin(2\pi l_mx/L+ \phi_m)$. The integration of the ground truth data is done with an implicit Runge-Kutta integration method of the Radau IIA family (order 5) \cite{brandstetter2022liepointsymmetrydata}, and our models use the RK4 solver with the same $\Delta t$. Finally, as for our previous experiments, we divide the training trajectories into non-overlapping chunks of $N=2\Delta t$ steps and evaluate the long-term stability on 128 test trajectories of duration $T_{\text{test}}=640\Delta t$. 

\subsection{Experimental setting}
Models are trained using Titan RTX GPU (cuda 13.2), each training taking approximately one hour to be achieved. The models for the short-online experiments are trained using one seed, while the models for the online experiments are trained using 3 different seeds. 

We use the MLP architecture from \cite{Yin_2021} for the Neural ODEs (i.e., for $F_\theta$). It is composed of three linear layers with activation ReLU and hidden size 200 for RB and KS, hidden size 64 for Van der Pol. All Neural ODEs use RK4 as integration scheme as well, and are trained using Adam optimizer \cite{kingma2017adammethodstochasticoptimization} with learning rate $l_r=0.001$, $\beta_1=0.9$ and $\beta_2=0.999$. For the $\mathcal{L}_{\text{Lip}}$ loss, we sample $V=10$ directions $\{v_i\}_{i=1}^V$ from $\mathcal{N}(0,I)$ at each epoch. Table \ref{tab:params} summarizes the parameters of the different models and experiments.

\begin{table}[h]
\caption{Summary of model and experiment parameters}
\label{tab:params}
  \begin{center}
    \begin{small}
      \begin{sc}
    \begin{tabular}{|l|c|c|c|}
    \hline
    \textbf{Parameters} & \textbf{Rigid Body} & \textbf{Kuramoto-Sivashinsky} & \textbf{Van der Pol} \\
    \hline
    $U$ dimension  & 3 & 256 & 2 \\
    \hline
    $\Delta t / \Delta t_{\text{model}}$ (s) & 0.01 / 0.1 & 0.2 / 0.2 & 0.01/ 0.1 or 0.01 \\
    \hline
    $T_{\text{train}} / T_{\text{val}} / T_{\text{test}}$ (s) & 15 / 15 / 800 & 28 / 28 / 128  & 10/10/10  \\
    \hline
    $N_{\text{train}} / N_{\text{val}} / N_{\text{test}}$ (s) & 5 / 5 / 8000 & 2 / 2 / 640 & 1/1/100 or 1000   \\
    \hline
    Nb traj train/val / test & 40 / 40 / 100 & 512 / 128 / 128 & 10/10/10  \\
    \hline
    Nb epochs & 4500 & 1000 & 10000 \\
    \hline
    Batch size & 40 & 128 & 100\\
    \hline
    Seeds & 1 & 1 & 3 \\ 
    \hline
    \end{tabular}
\end{sc}
\end{small}
\end{center}
\end{table}

\begin{figure}[h]
    \centering
    \includegraphics[width=0.45\linewidth]{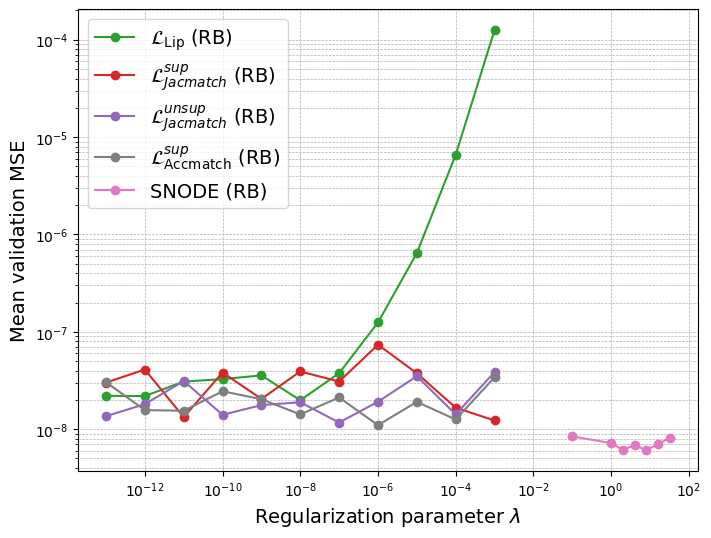}
    \includegraphics[width=0.45\linewidth]{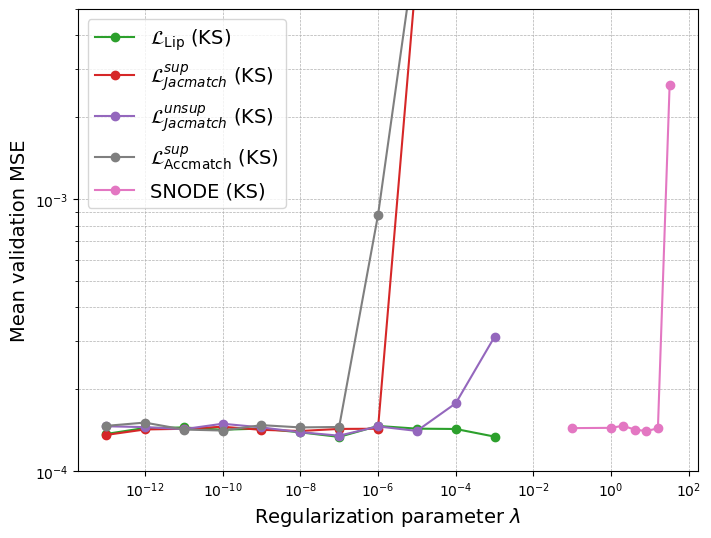}
    \caption{Mean MSE on validation set for the optimization of the $\lambda$ parameter, for the Rigid Body Problem (left) and KS (right) experiments.}
    \label{fig:val}
\end{figure}

The best model is determined based on the mean trajectory loss on a validation set of chunked trajectories. The optimal regularization coefficient $\lambda$ or $\gamma$ is determined by grid-search for each experiment, also based on the same metric for the last epoch, then take the best model for this metric of this $\lambda/\gamma$ run. Figure \ref{fig:val} shows its evolution across the tested hyperparameters $\lambda$, between 1E-2 to 1E-13 for RB and 1E-3 to 1E-13 for KS. The ranges of $\lambda$ for SNODE are the ones from White et al. \cite{white2024stabilizedneuraldifferentialequations}. 

\section{Additional experiment: Two-Body Problem}
\label{a:tb}
\paragraph{Two-Body Problem}
The Two-Body (TB) Problem describes the motion of two bodies attracting each other with a force inversely proportional to their squared distance, such as the gravitational force \cite{hairer_geometricnumericalintegration}. It is described by the following system of equations:
\begin{equation}
U=\begin{pmatrix}
    x \\
    y \\
    \dot{x} \\
    \dot{y}
\end{pmatrix}, \dot{U} = \begin{pmatrix}
    \dot{x} \\
    \dot{y} \\ 
    \frac{-x}{(x^2+y^2)^{3/2}} \\
    \frac{-y}{(x^2+y^2)^{3/2}}
\end{pmatrix}
\end{equation}
with a body fixed at the origin and $(x,y)$ the Cartesian coordinates of the other body in the orbit plane \cite{white2024stabilizedneuraldifferentialequations}. The angular momentum $L_0$ is conserved here: $x\dot{y}-\dot{x}y=L_0$ for $(x,y)\in \mathcal{T}_F^2$. As in White \cite{white2024stabilizedneuraldifferentialequations} experiments, we train on 40 trajectories with initial conditions $X_0=(1-e, 0, 0, \sqrt{\frac{1 + e}{1 - e}})$ with the eccentricity $e$ sampled from $\mathcal{U}(0.5,0.7)$. To reach a challenging context for Neural ODEs, we use data from trajectories of $T_{\text{train}}=8s$ which is less than a full period, integrated from $F_{\text{TB}}$ with a time-step $\Delta t=0.01s$. We then divide the training trajectories into non-overlapping chunks of $N=2\Delta t$ length. The ground truth and the models both use fourth order Runge-Kutta (RK4) with the same time-step $\Delta t$ as integration scheme to produce their trajectories. We then evaluate the long-term stability of our models on 100 test trajectories of duration $T_{\text{test}}=100s$. 

\begin{figure}
    \centering
    \includegraphics[width=0.45\linewidth]{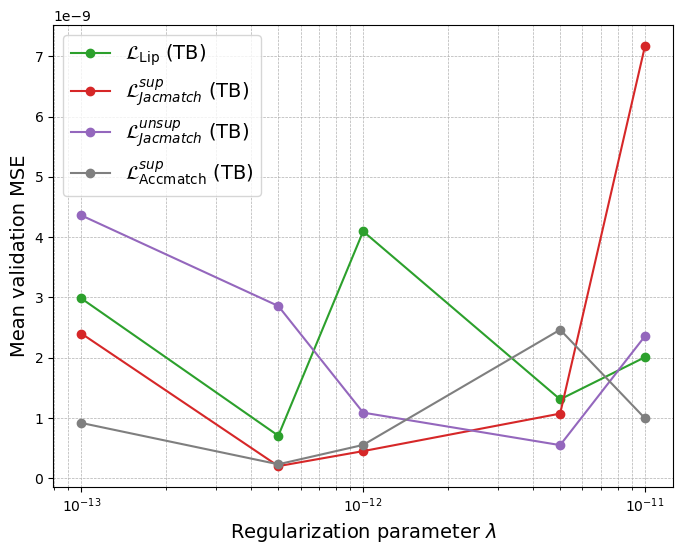}
    \includegraphics[width=0.45\linewidth]{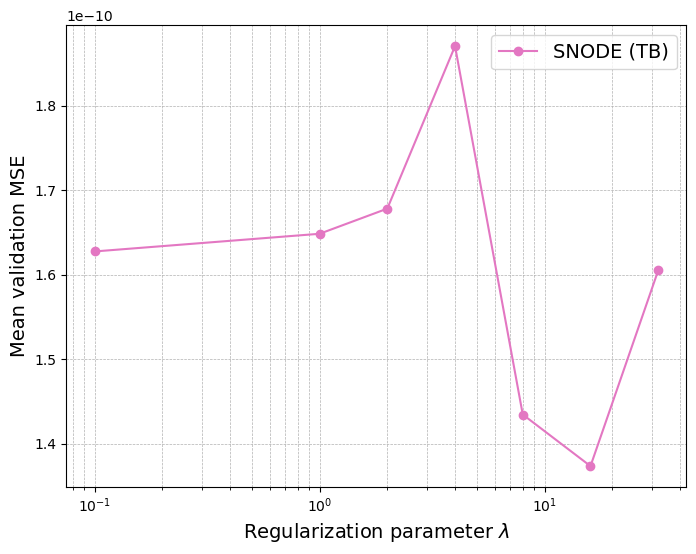}
    \caption{Mean MSE on validation set for the optimization of the $\lambda$ parameter, for the Two-Body Problem (Right: SNODE, Left: the other methods)}
    \label{fig:val-tb}
\end{figure}

Table \ref{tab:params-tb} summarizes the parameters of this experiment and Figure \ref{fig:val-tb} shows the results of the gridsearch performed for $\lambda$ in $\{1e-11, 5e-12, 1e-12, 5e-13, 1e-13 \}$.  

\begin{table}[h]
\caption{Summary of model and experiment parameters}
\label{tab:params-tb}
\begin{center}
\begin{small}
\begin{sc}
\begin{tabular}{|l|c|}
\hline
\textbf{Parameters} & \textbf{Two-Body} \\
\hline
$U$ dimension  & 4 \\
\hline
$\Delta t / \Delta t_{\text{model}}$ (s) & 0.01 / 0.01 \\
\hline
$T_{\text{train}} / T_{\text{val}} / T_{\text{test}}$ (s) & 8 / 8 / 100 \\
\hline
$N_{\text{train}} / N_{\text{val}} / N_{\text{test}}$ & 2 / 2 / 10000 \\
\hline
Nb traj train/val / test & 40 / 40 / 100 \\
\hline
Nb epochs & 4500 \\
\hline
Batch size & 40 \\
\hline
\end{tabular}
\end{sc}
\end{small}
\end{center}
\end{table}

\paragraph{Results} For this periodic system, all of the methods succeed in stabilizing the $N=2\Delta t$ baseline, with SNODE remaining the state-of-the-art model in this case, followed by the supervised JacMatch. The unsupervised version of $\mathcal{L}_{\text{Jacmatch}}$ only degrades a little bit the supervised version.

\begin{figure*}[t]  
  \centering
  \includegraphics[width=0.4\textwidth]{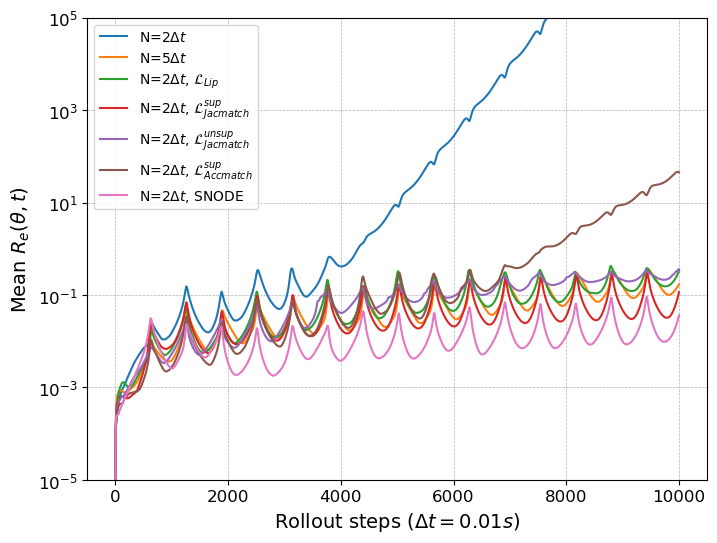}
  \includegraphics[width=0.59\textwidth]{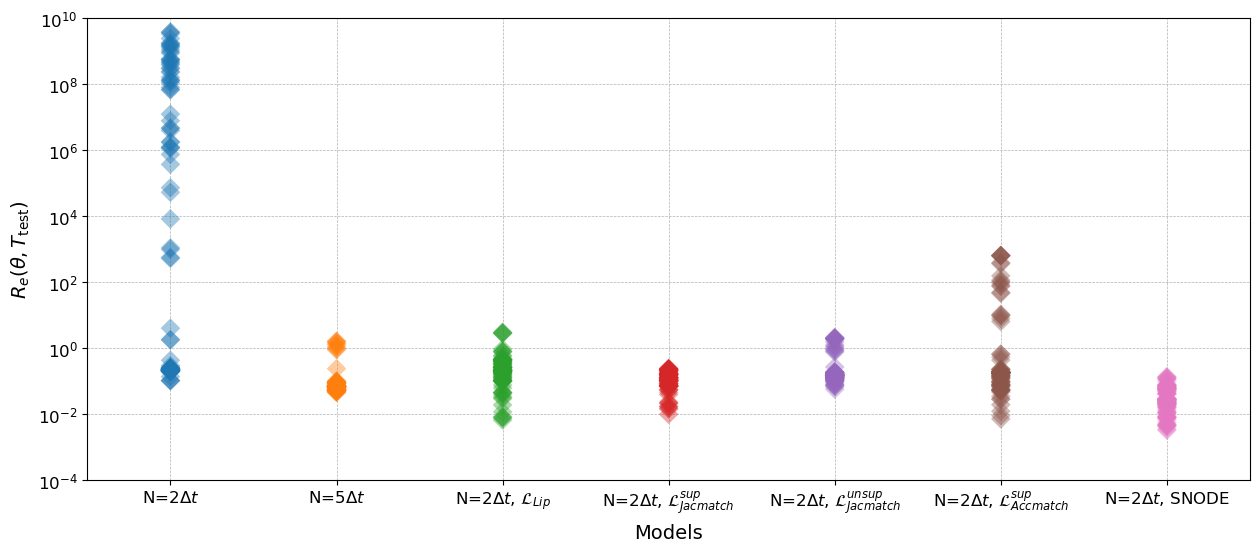}
    \includegraphics[width=0.4\textwidth]{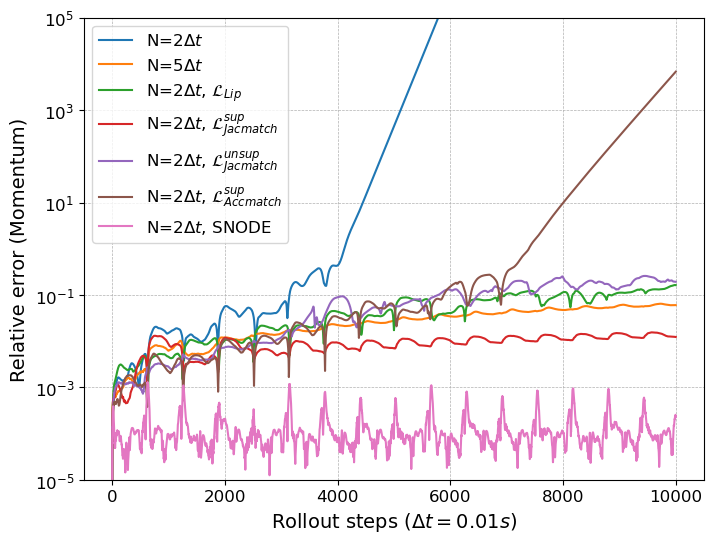}
      \includegraphics[width=0.59\textwidth]{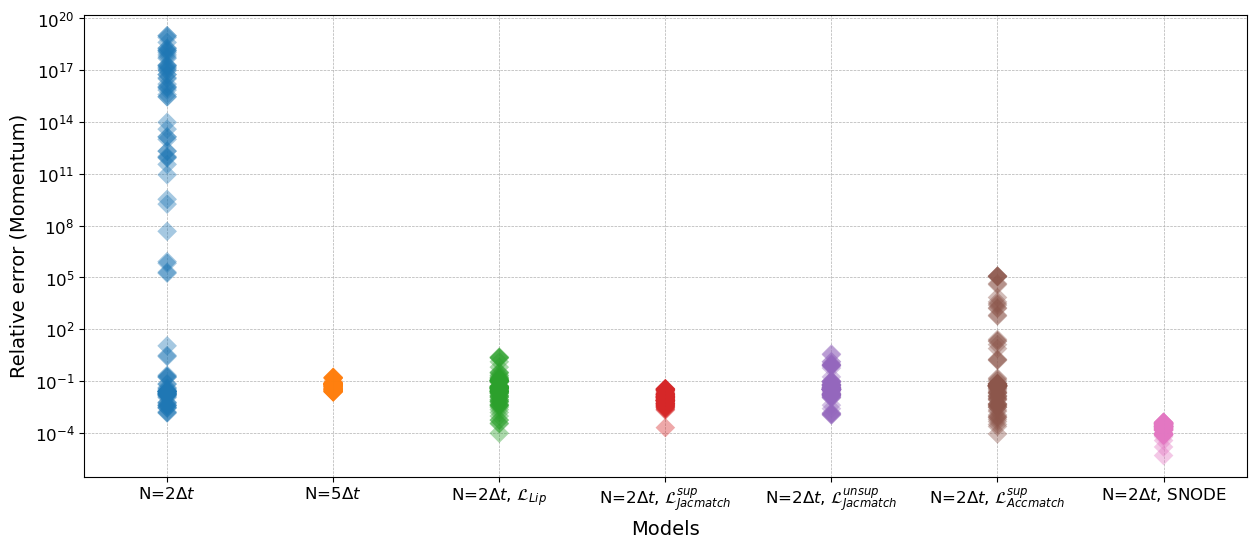}
  \caption{\textbf{Two Body.} For each row, we have the evolution of the relative error over time (left) and the distributions of final values (right). The top row represents the relative trajectory error $R_e(\theta,t)$ and the bottom row represents the relative conservation error (Momentum).}
  \label{fig:res-tb}
\end{figure*}

\section{Computational cost of methods}
\label{a:cost}
To compute the cost of our methods, we averaged the cost of single epochs over 1000 runs. We compare the regularized training losses for $N=2$ as in the main paper. $\mathcal{L}_\text{Lip}$ is computed with $V=10$. We also analyze the cost for larger-scale systems by testing of KS with dimensions 256 and 512. 

\begin{figure}
    \centering
    \includegraphics[width=0.5\linewidth]{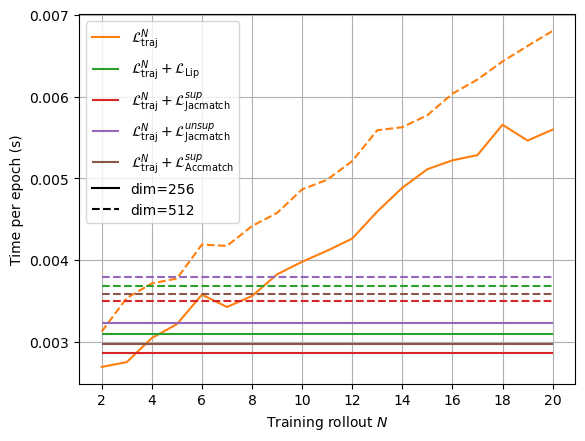}
    \caption{Computational cost on Kuramoto-Sivashinsky averaged over 1000 runs. For regularized methods, the cost for $N=2$ is shown to compare with the $N=10$ baseline.}
    \label{fig:cost}
\end{figure}

In Figure \ref{fig:cost}, we first observe the computational bottleneck of unrolling \cite{list2024differentiabilityunrolledtrainingneural}: the longer the training rollout, the larger the computational cost, and even more when going to higher dimensional systems. Being able to produce stable simulations when training on short rollouts appears very advantageous. 

Furthermore, all of the methods only introduce a small computational overhead, while remaining cheaper than training on $N=10$ (stable baseline), for both dimensions.

\section{Trajectories}
\begin{figure}[h]
    \centering
    \includegraphics[width=0.45\linewidth]{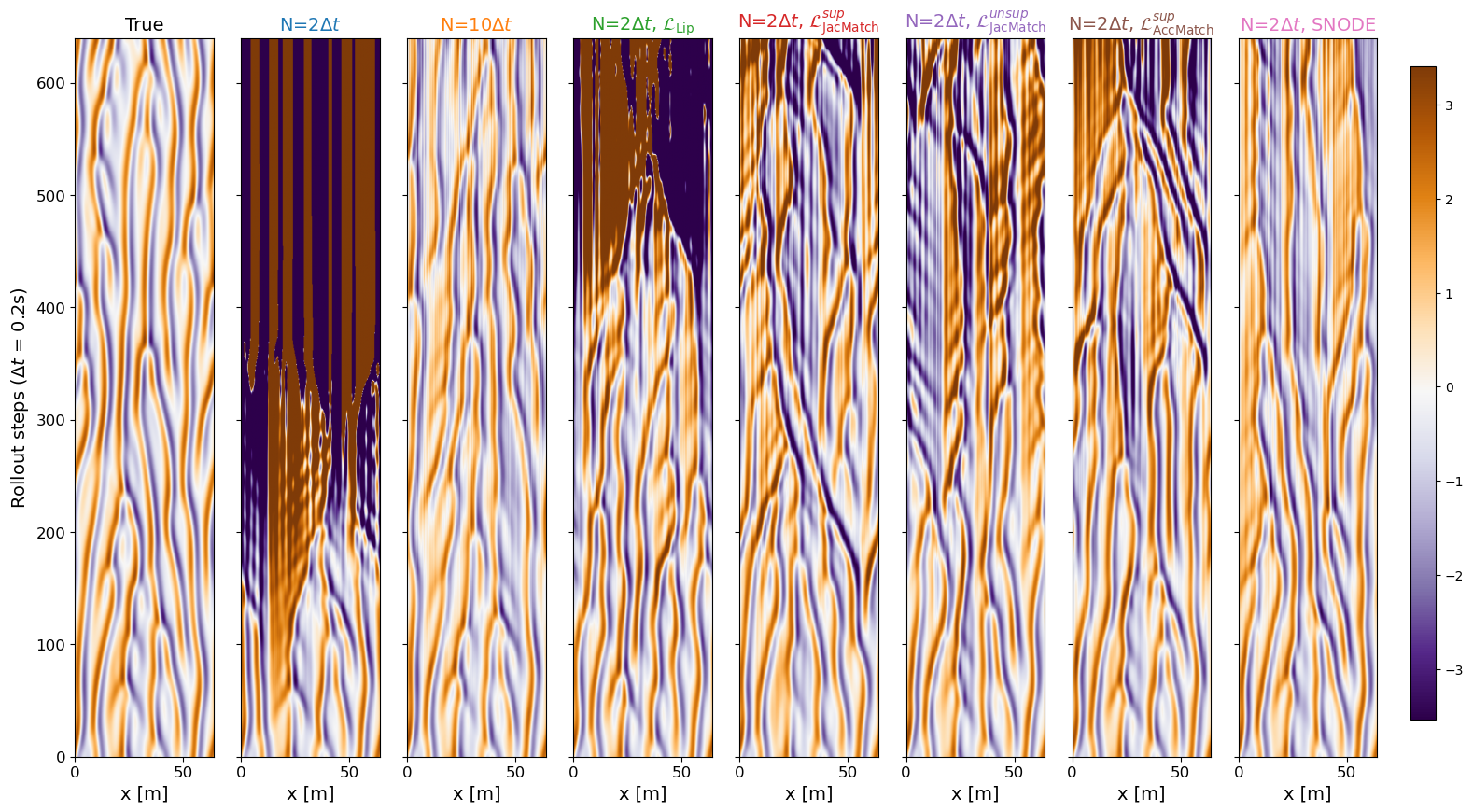}
    \includegraphics[width=0.45\linewidth]{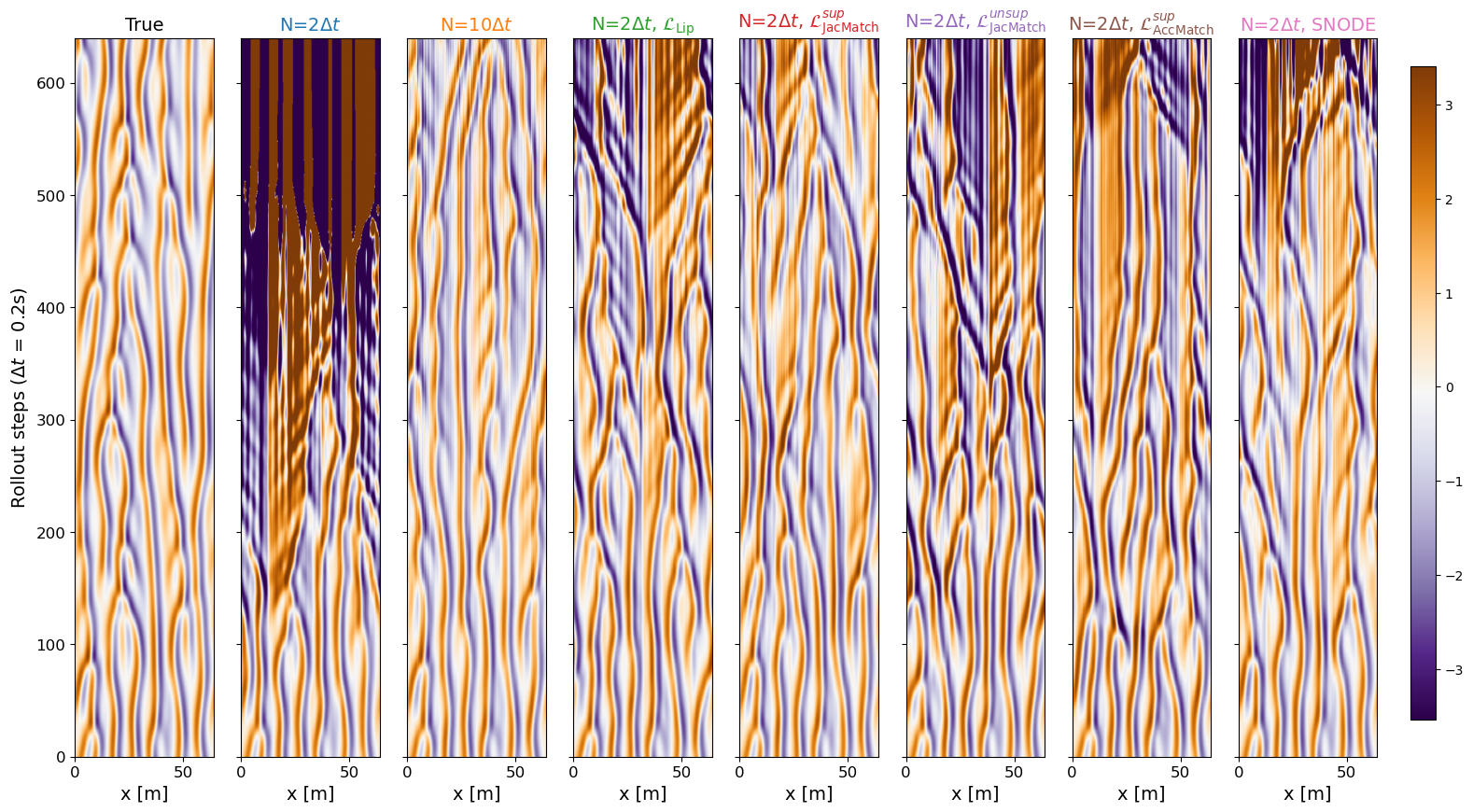}
    \includegraphics[width=0.45\linewidth]{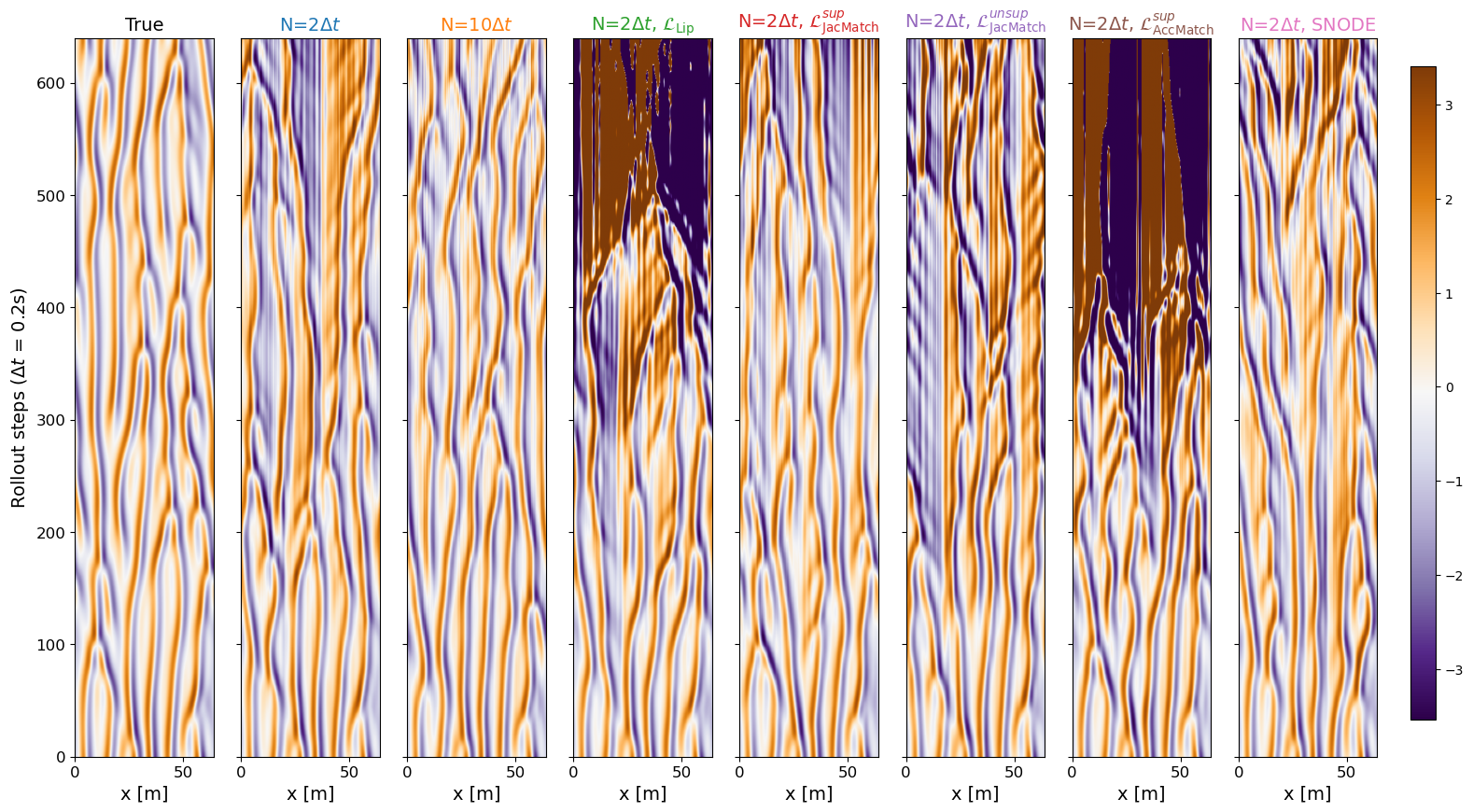}
    \includegraphics[width=0.45\linewidth]{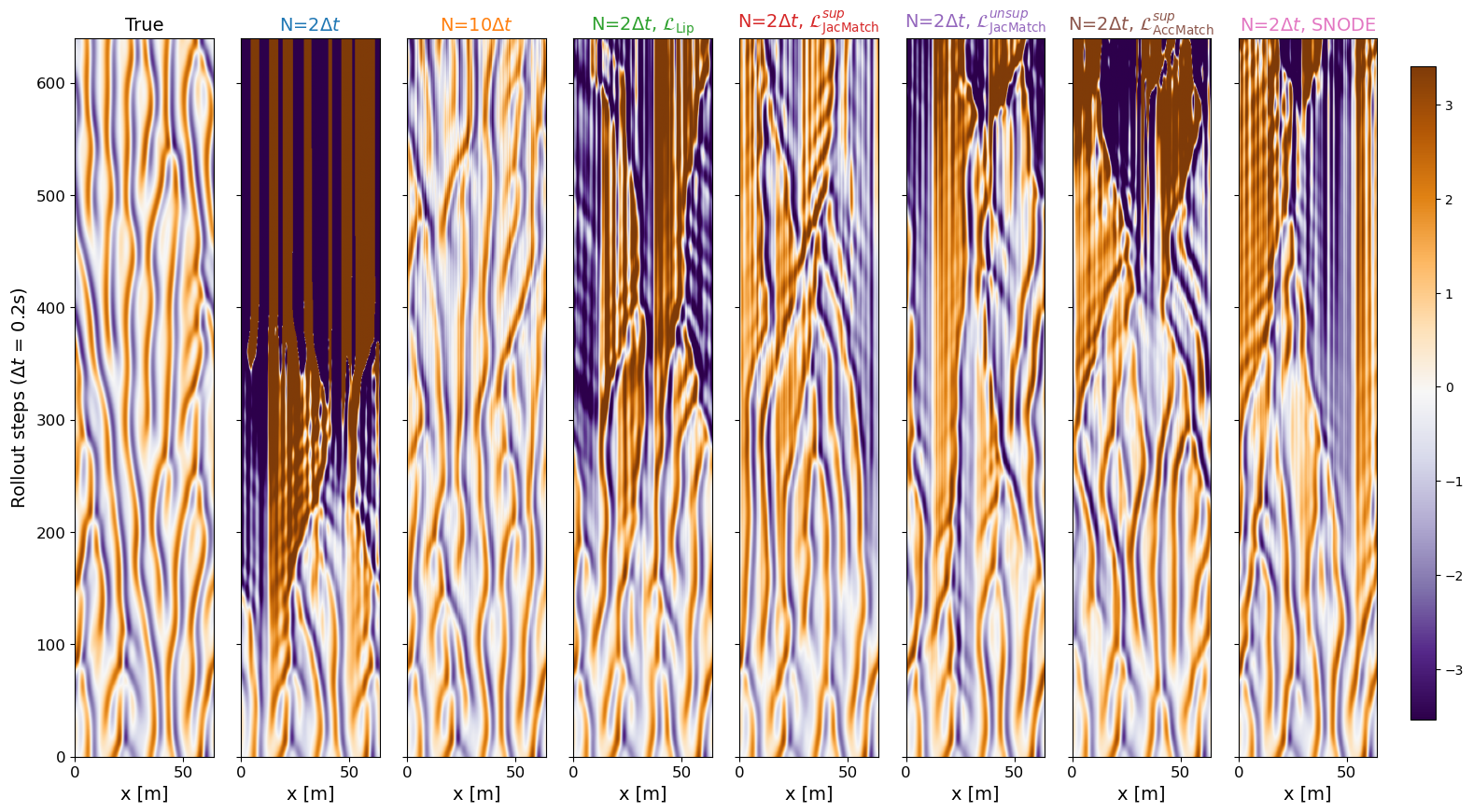}
    \includegraphics[width=0.45\linewidth]{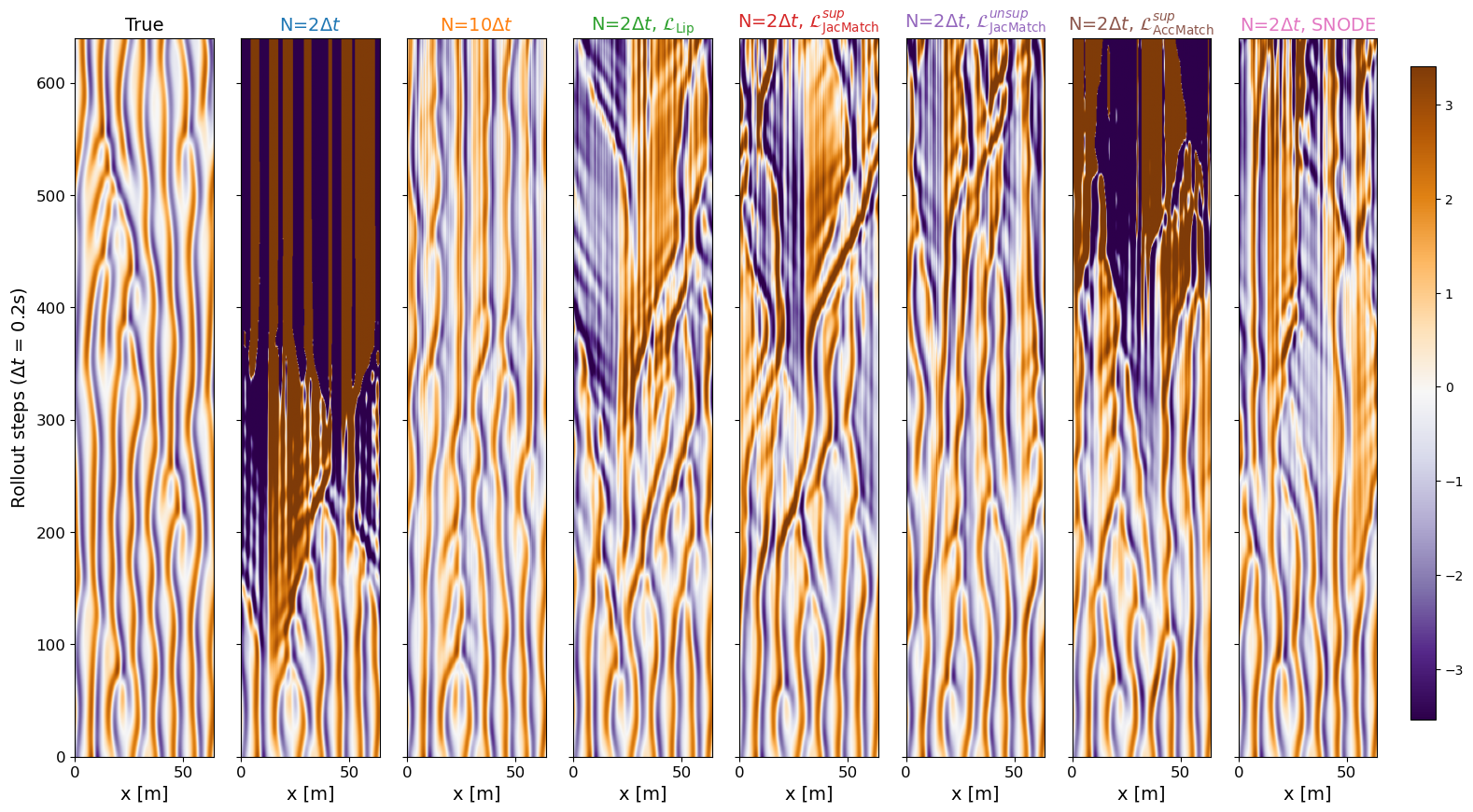}
    \includegraphics[width=0.45\linewidth]{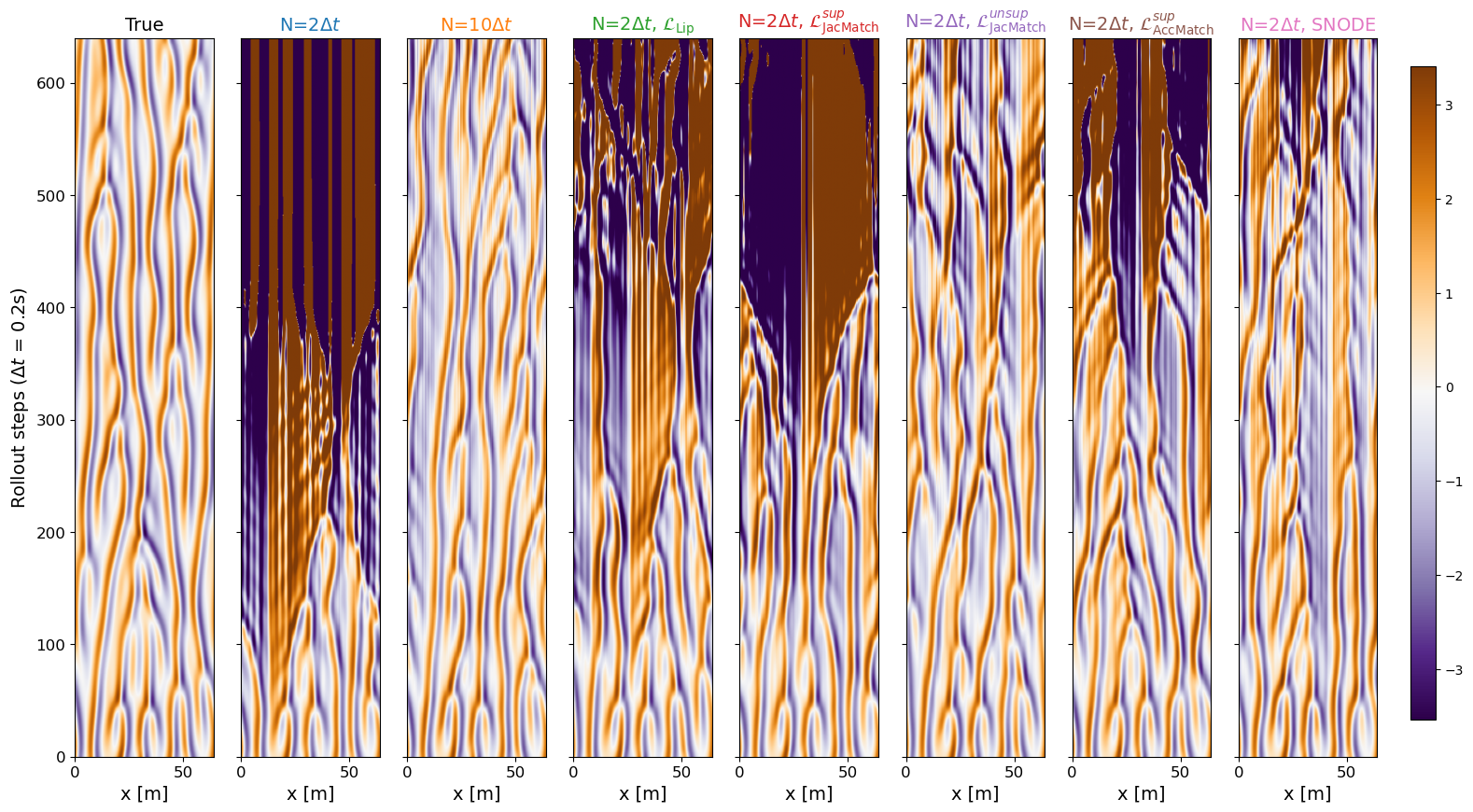}
    \caption{Kuramoto-Sivashinksy trajectories}
    \label{fig:ks_traj}
\end{figure}

\begin{figure}
    \centering
    \includegraphics[width=0.99\linewidth]{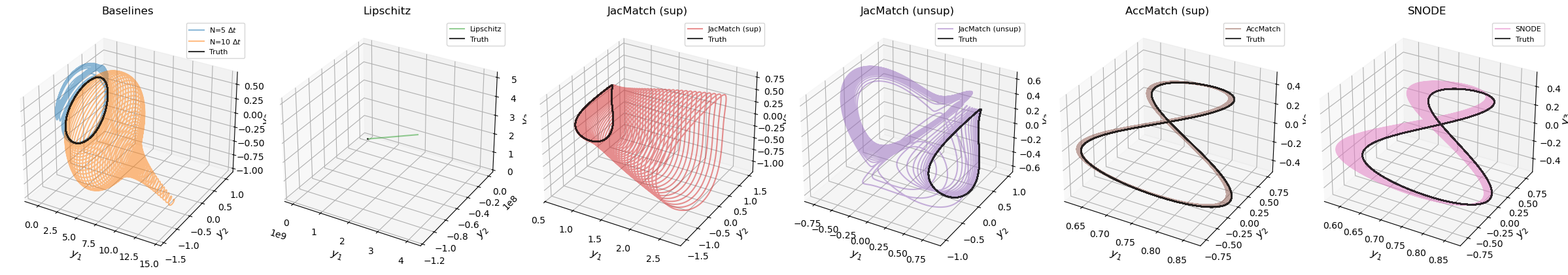}
    \includegraphics[width=0.99\linewidth]{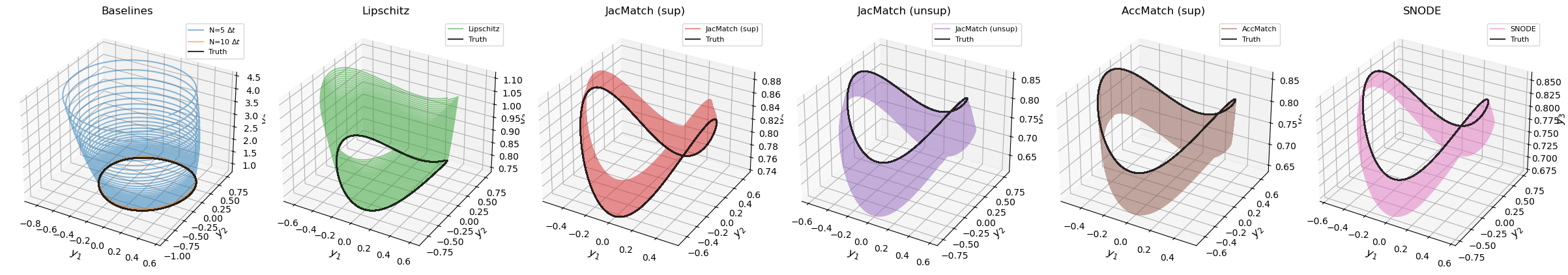}
    \includegraphics[width=0.99\linewidth]{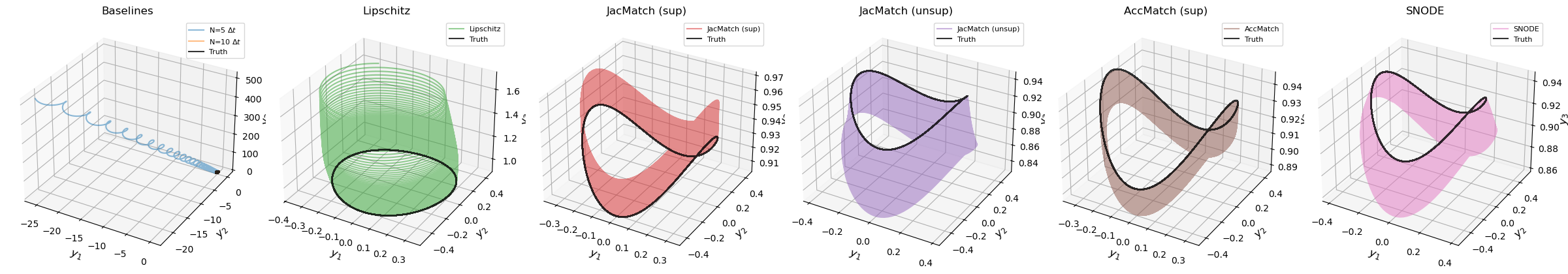}
    \includegraphics[width=0.99\linewidth]{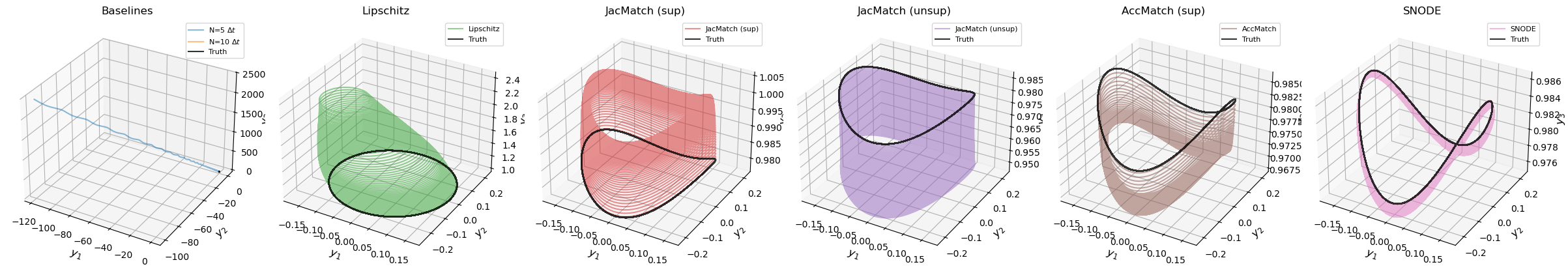}
    \includegraphics[width=0.99\linewidth]{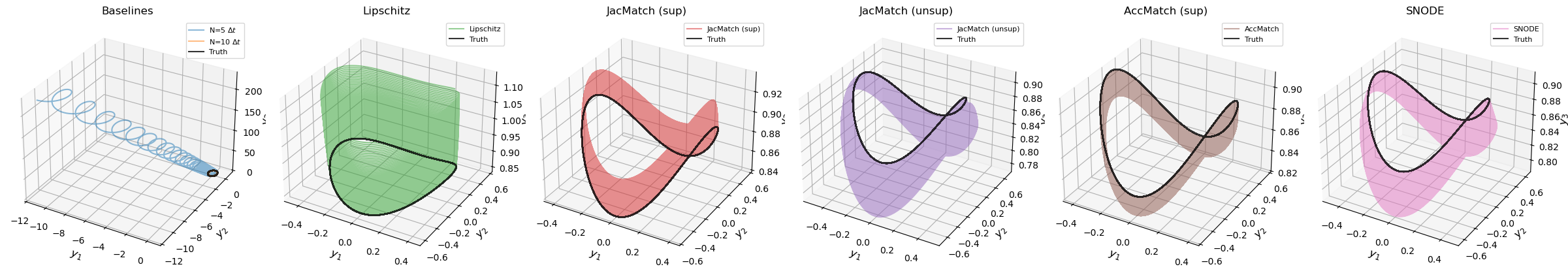}
    \caption{Rigid Body trajectories }
    \label{fig:rb_traj}
\end{figure}

We add some plots of the predicted trajectories to have a look at their qualitative aspect, in Figure \ref{fig:rb_traj} for Rigid Body and in Figure \ref{fig:ks_traj} for KS. 

\newpage 
 $\quad $

\newpage
\section*{NeurIPS Paper Checklist}
\begin{enumerate}

\item {\bf Claims}
    \item[] Question: Do the main claims made in the abstract and introduction accurately reflect the paper's contributions and scope?
    \item[] Answer: \answerYes{} 
    \item[] Justification: The claims made in the abstract and introduction reflect the paper contribution and scope, that are summarized in the conclusion \ref{sec:conclu}.  

\item {\bf Limitations}
    \item[] Question: Does the paper discuss the limitations of the work performed by the authors?
    \item[] Answer: \answerYes{} 
    \item[] Justification: The authors included the limitations of their work in the conclusion \ref{sec:conclu}, opening the door to future work. 

\item {\bf Theory assumptions and proofs}
    \item[] Question: For each theoretical result, does the paper provide the full set of assumptions and a complete (and correct) proof?
    \item[] Answer: \answerYes{} 
    \item[] Justification: All the theorems, formulas, and proofs in the paper are  numbered and cross-referenced. Every proof is detailed in the annexes and states the necessary assumptions to its application. Furthermore, the mathematical tools at the center of the proofs are mentioned in the main paper to provide intuition (e.g., "Grönwall's lemma", "Taylor-Young formula").

    \item {\bf Experimental result reproducibility}
    \item[] Question: Does the paper fully disclose all the information needed to reproduce the main experimental results of the paper to the extent that it affects the main claims and/or conclusions of the paper (regardless of whether the code and data are provided or not)?
    \item[] Answer: \answerYes{} 
    \item[] Justification: We provide the details of the experiments in both the main paper \ref{sec:results} and in annex \ref{a:implementation}, from the data generation protocols to the validation and testing setups or the architectures used. 

\item {\bf Open access to data and code}
    \item[] Question: Does the paper provide open access to the data and code, with sufficient instructions to faithfully reproduce the main experimental results, as described in supplemental material?
    \item[] Answer: \answerYes{} 
    \item[] Justification: We share our code in the submission to ensure reproducibility. The data can either be generated from our code or from other open-source codes (instructions to do so being detailed in annex \ref{a:implementation}). 

\item {\bf Experimental setting/details}
    \item[] Question: Does the paper specify all the training and test details (e.g., data splits, hyperparameters, how they were chosen, type of optimizer) necessary to understand the results?
    \item[] Answer: \answerYes{} 
    \item[] Justification: Training and test details are provided in annex \ref{a:implementation} along with the validation protocol we performed for our experiments.

\item {\bf Experiment statistical significance}
    \item[] Question: Does the paper report error bars suitably and correctly defined or other appropriate information about the statistical significance of the experiments?
    \item[] Answer: \answerYes{} 
    \item[] Justification:  Figures \ref{fig:res-rb} and \ref{fig:res-ks} report the distribution of final metric values over 100/128 test trajectories with diverse initial conditions (right panels), which captures the variance induced by sensitivity to initial conditions, the primary source of variability in chaotic systems.
    

\item {\bf Experiments compute resources}
    \item[] Question: For each experiment, does the paper provide sufficient information on the computer resources (type of compute workers, memory, time of execution) needed to reproduce the experiments?
    \item[] Answer: \answerYes{} 
    \item[] Justification: The type of computer resources used and the mean runtime of the experiments are mentioned in annex \ref{a:implementation}. 
    
\item {\bf Code of ethics}
    \item[] Question: Does the research conducted in the paper conform, in every respect, with the NeurIPS Code of Ethics \url{https://neurips.cc/public/EthicsGuidelines}?
    \item[] Answer: \answerYes{} 
    \item[] Justification: The paper follows the NeurIPS Code of Ethics. 

\item {\bf Broader impacts}
    \item[] Question: Does the paper discuss both potential positive societal impacts and negative societal impacts of the work performed?
    \item[] Answer: \answerNA{} 
    \item[] Justification: The scope of this work is mostly tied to climate or other physical modeling. 
    
\item {\bf Safeguards}
    \item[] Question: Does the paper describe safeguards that have been put in place for responsible release of data or models that have a high risk for misuse (e.g., pre-trained language models, image generators, or scraped datasets)?
    \item[] Answer: \answerNA{} 
    \item[] Justification: The paper does not pose any risk for misuse and does not release nor use sensitive data.

\item {\bf Licenses for existing assets}
    \item[] Question: Are the creators or original owners of assets (e.g., code, data, models), used in the paper, properly credited and are the license and terms of use explicitly mentioned and properly respected?
    \item[] Answer: \answerYes{} 
    \item[] Justification: Usage of existing assets is limited to open-source code from GitHub for generating data. Authors are cited. 

\item {\bf New assets}
    \item[] Question: Are new assets introduced in the paper well documented and is the documentation provided alongside the assets
    \item[] Answer: \answerNA{} 
    \item[] Justification: The paper does not release new assets apart from the code available on GitHub. 

\item {\bf Crowdsourcing and research with human subjects}
    \item[] Question: For crowdsourcing experiments and research with human subjects, does the paper include the full text of instructions given to participants and screenshots, if applicable, as well as details about compensation (if any)? 
    \item[] Answer: \answerNA{} 
    \item[] Justification: The paper does not involve crowdsourcing nor research with human subjects.

\item {\bf Institutional review board (IRB) approvals or equivalent for research with human subjects}
    \item[] Question: Does the paper describe potential risks incurred by study participants, whether such risks were disclosed to the subjects, and whether Institutional Review Board (IRB) approvals (or an equivalent approval/review based on the requirements of your country or institution) were obtained?
    \item[] Answer: \answerNA{} 
    \item[] Justification: The paper does not involve crowdsourcing nor research with human subjects.

\item {\bf Declaration of LLM usage}
    \item[] Question: Does the paper describe the usage of LLMs if it is an important, original, or non-standard component of the core methods in this research? Note that if the LLM is used only for writing, editing, or formatting purposes and does \emph{not} impact the core methodology, scientific rigor, or originality of the research, declaration is not required.
    \item[] Answer: \answerNA{} 
    \item[] Justification: LLMs are not important, original or non-standard components of this specific paper.

\end{enumerate}

\end{document}